\documentclass[11pt]{article}

\usepackage[final]{acl}

\usepackage{times}
\usepackage{latexsym}
\usepackage{booktabs}
\usepackage{amsmath}
\usepackage[T1]{fontenc}

\usepackage[utf8]{inputenc}

\usepackage{microtype}

\usepackage{inconsolata}

\usepackage[most]{tcolorbox}
\usepackage{enumitem}
\usepackage{amssymb}
\usepackage{graphicx}

\title{What Makes a Question Curious? \\ Measuring Model Curiosity in Human-LLM Interactions}

\author{Angana Borah$^{*1}$\hspace{5pt} 
Rachana Ramchandar$^{*2}$\hspace{5pt} 
Inderjeet Nair$^1$\hspace{5pt} 
 Rada Mihalcea$^1$  \\
$^1$University of Michigan - Ann Arbor, USA  \\
$^2$PES University, Bangalore, India \\
\textit{\{anganab, inair, mihalcea\}@umich.edu} \\ 
\textit{rachramchandar10@gmail.com} \\ }

\begin{document}
\maketitle
\begin{abstract}
Large language models increasingly act as partners in everyday communication. In these interactions, useful assistance may require not only answering, but also asking the right questions that surface knowledge gaps, assumptions, and alternatives. In this study, we refer to such epistemic question-asking as \textit{model curiosity}, and introduce a theory-grounded taxonomy that distinguishes curiosity-oriented questions from generic model questioning. Across four conversation datasets, we find that curiosity-oriented questions remain rare (17-28\% of all questions), but are associated with more exploratory and epistemically rich subsequent user responses, with varying patterns across datasets. We further conduct a human study to evaluate the usefulness of model curiosity, finding that curiosity can encourage exploration, especially in advice, creative, and research contexts. These findings have implications for building LLMs that ask more selectively to support deeper exploration and human-AI collaboration. We publicly release our code\footnote{\url{https://github.com/MichiganNLP/curiosity_importance}} and website\footnote{\url{https://lit.eecs.umich.edu/curiosity_importance/}} to support reproducibility and future research.






\end{abstract}

\section{Introduction}

Large language models are extensively used today as partners in learning, writing, advice-seeking and collaboration~\cite{chatterji2025people, mysore2025prototypical}. As tasks become more open-ended, useful assistance is no longer limited to producing an answer; it can also involve helping users reflect on goals and assumptions~\citep{park2023thinking}. This is tied to the concept of \textit{curiosity}: an orientation toward noticing gaps or possibilities in one's understanding and seeking information, perspectives, or alternatives to resolve or explore them~\cite{loewenstein1994psychology, litman2005curiosity, kidd2015psychology}. In human interactions, curiosity plays a central role by driving deeper exploration~\cite{sinha2017new, dubey2021contagious}. In human-LLM interactions, model curiosity may play a similar role by shaping whether users simply receive information or engage more deeply.


\begin{figure}
\centering
\includegraphics[width=\linewidth]{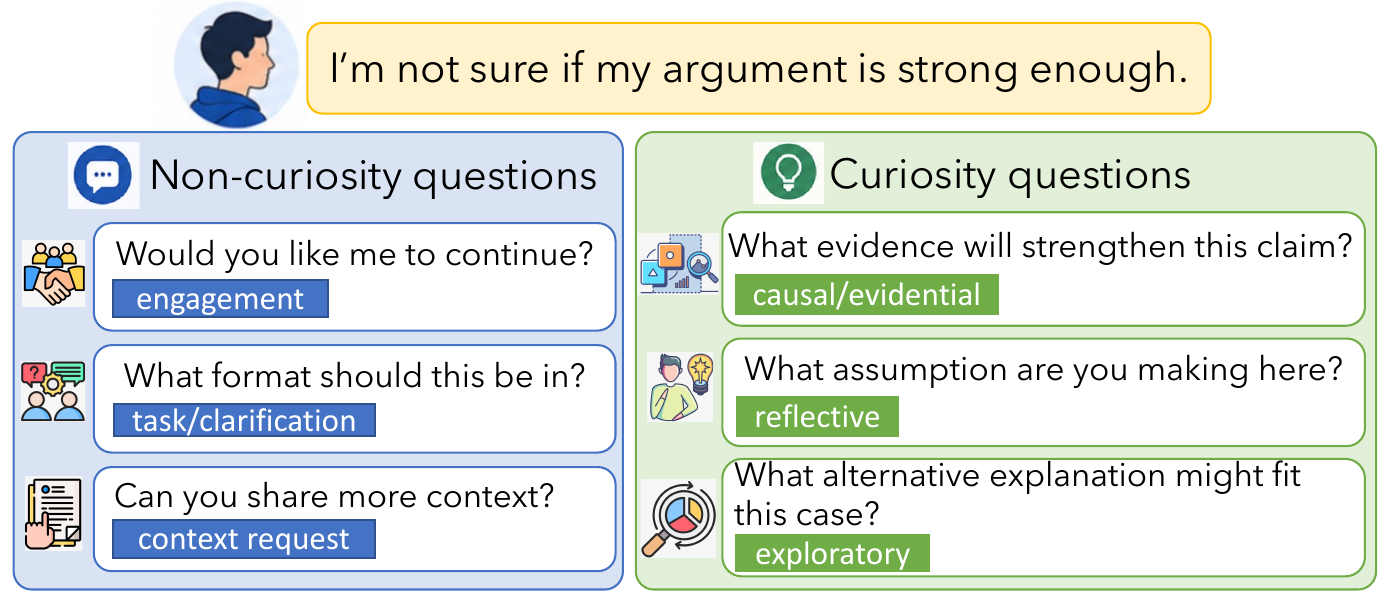}
\caption{\textbf{Overview of questions in human-model interaction.} Given the same user prompt, we define a taxonomy to distinguish between the two types: (1) non-curiosity questions support task flow, clarification, or engagement, and (2) curiosity-oriented questions invite evidence-seeking, reflection, and exploration instead of directly answering with assumptions. Note that each category includes additional subtypes, we show three here for illustration.}
\vspace{-10.5pt}
\label{fig:curiosity_main}
\end{figure}

Despite this potential, curiosity remains under-investigated as a conversational behavior in LLMs. Closest existing work on LLM interaction focuses on instruction following~\cite{zhou2023instruction}, clarification~\cite{aliannejadi2019asking}, proactive dialogue~\cite{deng2023prompting}, or personalized question generation~\cite{bowden2024active}. Work on LLM curiosity often operationalizes it through question generation~\citep{javaji2024would,iji2025chain, borah-etal-2025-curious}, reflecting a broader tradition of using question asking as a curiosity proxy~\citep{berlyne1954experimental, berlyne1966curiosity}. But model questions can serve different functions: some clarify task requirements or maintain engagement, while others invite users to examine evidence, assumptions, alternatives, or uncertainty (example in Fig~\ref{fig:curiosity_main}). We refer to the latter as curiosity-oriented: questions that may support deeper exploration and collaboration in human-LLM interaction.
  
In this study, we ask three research questions: \textbf{RQ1:} What kind of questions do LLMs ask in the wild, and how often are they curiosity-oriented? \textbf{RQ2:} Are curiosity-oriented model questions associated with richer downstream user engagement and curiosity? \textbf{RQ3:} When is model curiosity useful rather than unnecessary/burdensome? It is also crucial to note that we use \textit{model curiosity} to describe observable epistemic behavior, without attributing an internal psychological state to LLMs.




To address the above questions, we summarize our contributions as follows: (1) For RQ1, we introduce a taxonomy that distinguishes curiosity-oriented LLM questions from generic questions and use it to analyze model behavior in real human-LLM conversations. (2) For RQ2, we analyze whether model curiosity is associated with downstream user engagement and curiosity, finding that curiosity-oriented questions are followed by richer engagement and curiosity. Finally, (3) for RQ3, we introduce a human-validated framework for useful model curiosity, showing that usefulness is context-dependent. We also provide a proof-of-concept model for predicting when curiosity-oriented questions are likely to be useful. Together, our work defines model curiosity and highlights its role in human-LLM interaction.

\section{Related Work}

Curiosity is a fundamental driver of inquiry and learning: it directs attention to knowledge gaps and motivates information seeking~\cite{loewenstein1994psychology}. Psychological work identifies novelty, complexity, and ambiguity as common triggers for curiosity~\cite{berlyne1966curiosity,litman2010validity}. NLP research has primarily examined human question asking, clarification, and personalized question generation~\cite{rao-daume-iii-2018-learning,aliannejadi2019asking,ceraolo-etal-2025-quriosity}. Recent work studies cultural variation in curiosity across humans and models~\cite{borah-etal-2025-curious}. However, model-level curiosity remains underdefined and insufficiently grounded in prior theories, making it unclear which model questions should count as curiosity-oriented  in natural interaction.

Previous work shows that socratic questioning can promote reflection~\cite{ng2024socratic}, and that model-generated reflective questions can enhance decision-making~\cite{park2023thinking}. Furthermore, curiosity can spread through influence in human societies~\cite{dubey2021contagious}, consistent with broader contagion mechanisms~\cite{hatfield1993emotional}. Therefore, investigating whether curiosity in LLM questions is associated with greater user curiosity and engagement helps show why model curiosity may be important in interactions.



LLM questions serve different interactional functions: they can resolve ambiguity, specify tasks, maintain engagement, or invite reflection~\citep{aliannejadi2019asking,deng-etal-2023-prompting,bowden-etal-2024-active}. Because such questions can shape user agency, model curiosity should be selective rather than indiscriminate: some questions support reflection and exploration, while others may feel burdensome, intrusive, or steering~\citep{amershi2019guidelines,shneiderman2020human}. The distinction matters because models can steer conversational direction and depth \cite{yun2026agenda}, while preference optimization may encourage agreeable or sycophantic behavior~\cite{sharma2023sycophancy}. This motivates our taxonomy, which distinguishes  curiosity from generic questioning, and analysis of when model curiosity is genuinely useful.


\begin{table*}[ht]
\centering
\footnotesize
\setlength{\tabcolsep}{5pt}
\renewcommand{\arraystretch}{1.15}
\begin{tabular}{@{}p{0.28\textwidth}p{0.40\textwidth}p{0.25\textwidth}@{}}
\toprule
\textbf{Question type} & \textbf{Theoretical grounding} & \textbf{Example} \\

\midrule

Curiosity: Information-gap
& Information-gap theory; epistemic curiosity
& ``What problem does it solve?'' \\

Curiosity: Causal-explanatory
& Explanation-seeking; epistemic curiosity; explanations as understanding
& ``Why did this happen?'' \\

Curiosity: Exploratory broadening
& Epistemic exploration; diversive and specific curiosity
& ``What else might explain this?'' \\

Curiosity: Reflective-metacognitive
& Metacognition; epistemic reflection; uncertainty monitoring
& ``What assumptions are you making?'' \\

Curiosity: Perspective-taking
& Perspective-taking; role-taking theory; empathy
& ``How might they interpret this?'' \\

Curiosity: Counterfactual-hypothetical
& Counterfactual reasoning; hypothetical exploration; epistemic simulation
& ``What if the context changed?'' \\

\midrule

Non-curiosity: Context request
& Grounding in communication; task-oriented information gathering
& ``What topic should this be about?'' \\

Non-curiosity: Clarification repair
& Conversation repair; grounding; ambiguity resolution
& ``What do you mean by X?'' \\

Non-curiosity: Task management
& Task-oriented dialogue; conversational grounding
& ``What format should I use?'' \\

Non-curiosity: Engagement maintenance
& Engagement management; continuation behavior
& ``Would you like me to continue?'' \\

\midrule

Problematic: Privacy probing
& Privacy-sensitive information seeking; safety risks
& ``Can you share personal details?'' \\

Problematic: Leading or steering
& Leading questions; biased or steering interaction
& ``Isn't this clearly best?'' \\



\bottomrule
\end{tabular}
\caption{Taxonomy of assistant-to-user questions. Curiosity-oriented questions are grounded in theories of information gaps, explanation-seeking, epistemic exploration, metacognition, perspective-taking, and counterfactual reasoning.}
\vspace{-15pt}
\label{tab:curiosity-taxonomy}
\end{table*}

\section{Curiosity Data and Taxonomy}

Curiosity is defined as an \textit{epistemic behavior}: surfacing knowledge gaps, seeking explanations, examining assumptions, and considering alternatives~\cite{loewenstein1994psychology}. However, in human-LLM interactions, LLMs ask questions for many reasons, from ambiguity resolution to maintaining engagement and prompting exploration. Without separating these functions, an engagement prompt `Would you like me to continue?'' would be treated the same as a curiosity-oriented one, `What assumption are you making here?'' So, we develop a taxonomy of LLM questions in real interactions, distinguishing curiosity-oriented questions from non-curiosity questions to evaluate LLM behavior.

\subsection{Data}
We analyze four public conversation datasets: WildChat~\citep{zhao2024wildchat}, OASST1~\citep{kopf2023openassistant}, LMSYS-Chat-1M~\citep{zheng2024lmsys}, and ShareGPT\footnote{\url{https://sharegpt.com/}}. Together, they cover diverse human-LLM interactions. 
For each dataset, we extract user queries, model questions, and all subsequent user responses available. This produces a shared event-level and trajectory-level representation across datasets, allowing us to analyze questions models ask and how users respond to them. Dataset details and preprocessing are provided in Appendix~\ref{app:data}. 

It is also important to note that unlike the other corpora, OASST1 contains human-authored
assistant-role responses rather than naturally occurring
model-generated replies. Therefore, we treat this dataset as a comparison corpus for assistant-style questioning and
analyze the downstream engagement separately from the human-LLM datasets.

\subsection{Taxonomy Construction} 

Our taxonomy separates curiosity-oriented questions from non-curiosity (generic) questions, grounding in theories of information-gap and epistemic curiosity~\cite{loewenstein1994psychology, litman2005curiosity}, self-explanation in learning~\cite{chi1994eliciting}, mixed-initiative tutoring dialogue~\cite{graesser2005autotutor} and so on. This leads to six dimensions for curiosity questions: (1) \textit{information-gap} (surface missing epistemic knowledge), (2) \textit{causal-explanatory} (reasons/mechanisms, why/how something happens), (3) \textit{exploratory-broadening} (opens additional interpretations/implications/directions), (4) \textit{reflective-metacognitive}(ask user to examine own assumptions/evidence/uncertainty), (5) \textit{perspective-taking} (invites consideration of another person or viewpoint), and (6) \textit{counterfactual-hypothetical}: (how a situation would change under imagined/altered conditions). We provide theory and subtype details in Appendices~\ref{app:theory} and~\ref{app:taxonomy_extended}, and boundary-case examples distinguishing close categories in Appendix~\ref{app:boundary_case}.

We also include non-curiosity categories drawing from work on clarification in conversational systems~\cite{rahmani2023survey} and task-oriented information-seeking dialogue~\cite{feng2023towards}. They include: (1) \textit{context requests} (task information to complete the request), (2) \textit{clarification repair} (resolving ambiguity), (3) \textit{task management} (on format/tone/language/tool use) and (4) \textit{engagement maintenance} (conversation continuation prompts~\cite{sidner2005explorations}). Unlike curiosity-oriented questions, these questions help LLMs proceed with the task or manage the interaction. We also separate potentially problematic questions, including \textit{privacy probing} and \textit{leading or steering}. We remove all artifacts such as roleplay dialogue, interview templates, or malformed fragments that contain question marks but are not LLM-to-user questions. Table~\ref{tab:curiosity-taxonomy} illustrates examples for each curiosity and non-curiosity-subtypes. 

We operationalize this taxonomy in three stages: \textbf{(1) Rule-based pre-processing and extraction}: identify candidate LLM questions and filter obvious artifacts, such as generated dialogue, quizzes, templates, code blocks, user-requested question and so on. \textbf{(2) LLM-judge classification}: For each candidate question, \texttt{gpt-5-mini} receives
the preceding user turn, full assistant response, and
extracted question, together with the taxonomy, boundary
rules, and examples. The judge first determines whether the candidate is a genuine assistant-to-user question, then assigns a top-level status (curiosity, non-curiosity, problematic, or unclear) and a question subtype.
We use categorical labels rather than independent scores for each class. Further details are provided in Appendix~\ref{app:llmjudge}.
Finally, \textbf{(3) Human Validation}: validate judge annotations on 300 examples (260 unique examples and 20 shared items). The 20 shared items were sampled across all four datasets. They cover curiosity, non-curiosity, problematic, unclear and generated-artifact cases. This shared subset is used
to assess inter-annotator consistency, rather than to
estimate dataset-specific reliability. Three annotators label whether the extracted question is real, curiosity-oriented, and, if so, its subtype. Further details on preprocessing, LLM Judge Prompt and human validation is provided in Appendix~\ref{app:llmjudge}.







\begin{figure}
\centering
\includegraphics[width=\linewidth]{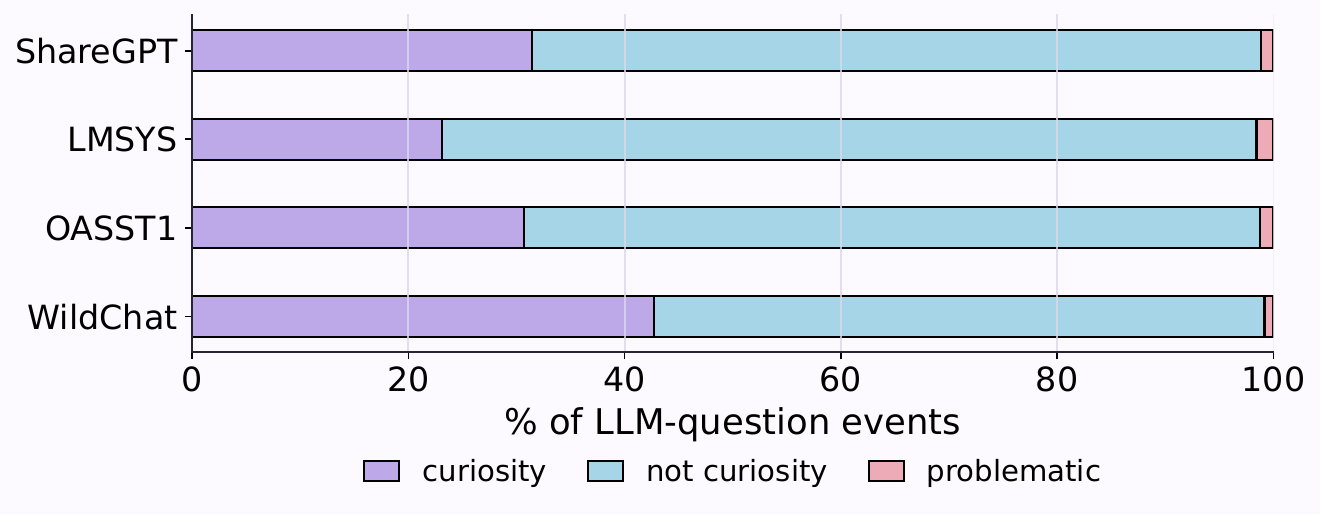}
\caption{Labels across datasets. Curiosity-oriented questions appear less frequently, with WildChat showing the highest share.}
\vspace{-10.5pt}
\label{fig:stacked_full}
\end{figure}

\section{How do LLMs Ask Questions in the Wild?}


We apply the above taxonomy to classify curiosity/non-curiosity/problematic questions. Fig~\ref{fig:stacked_full} shows that across datasets, curiosity-oriented questions are present but rare. Many model questions are instrumental or engagement-preserving rather than curious. For example, in WildChat, context requests and engagement-maintenance questions are among the largest non-curiosity categories. While such questions can be useful, they do not necessarily prompt reasoning, exploration, or reflection. This highlights the need to distinguish curiosity-question from non-curiosity questions.


\begin{figure}
\centering
\includegraphics[width=\linewidth]{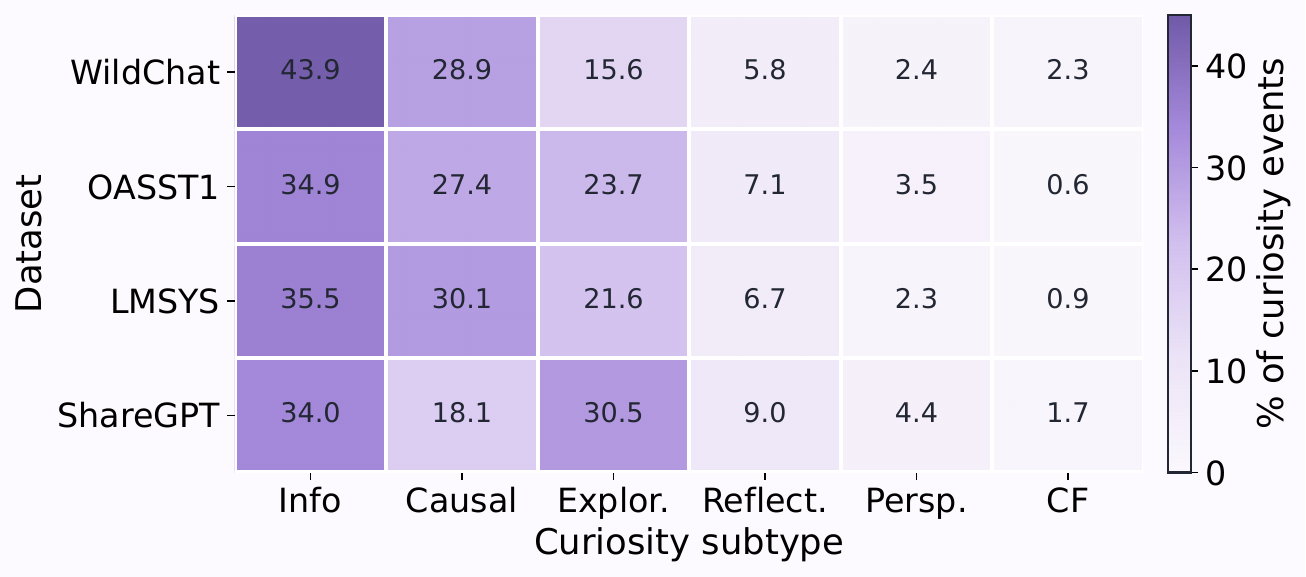}
\caption{Curiosity across datasets}
\vspace{-10.5pt}
\label{fig:cur_compare_data}
\end{figure}

\noindent \textbf{Curiosity type Distribution.} Curiosity is also unevenly distributed across types. Fig~\ref{fig:cur_compare_data} show that across datasets, models most often ask \textit{information-gap}, \textit{causal}, and \textit{exploratory} questions. While \textit{reflective}, \textit{perspective-taking}, and \textit{counterfactual} questions are consistently rarer. Overall, LLMs show some curiosity-like behavior, but rarely to support deeper reflection or sustained inquiry, motivating the need for methods that encourage useful forms of model curiosity.

 


\noindent \textbf{Topic/Antecedent Analysis.} 
We examine which user contexts elicit curiosity-oriented LLM questions. For this, we label the preceding user turn by topic and request form, then measure whether the following LLM question is curiosity-oriented.

\begin{figure}[h]
\centering
\includegraphics[width=\linewidth]{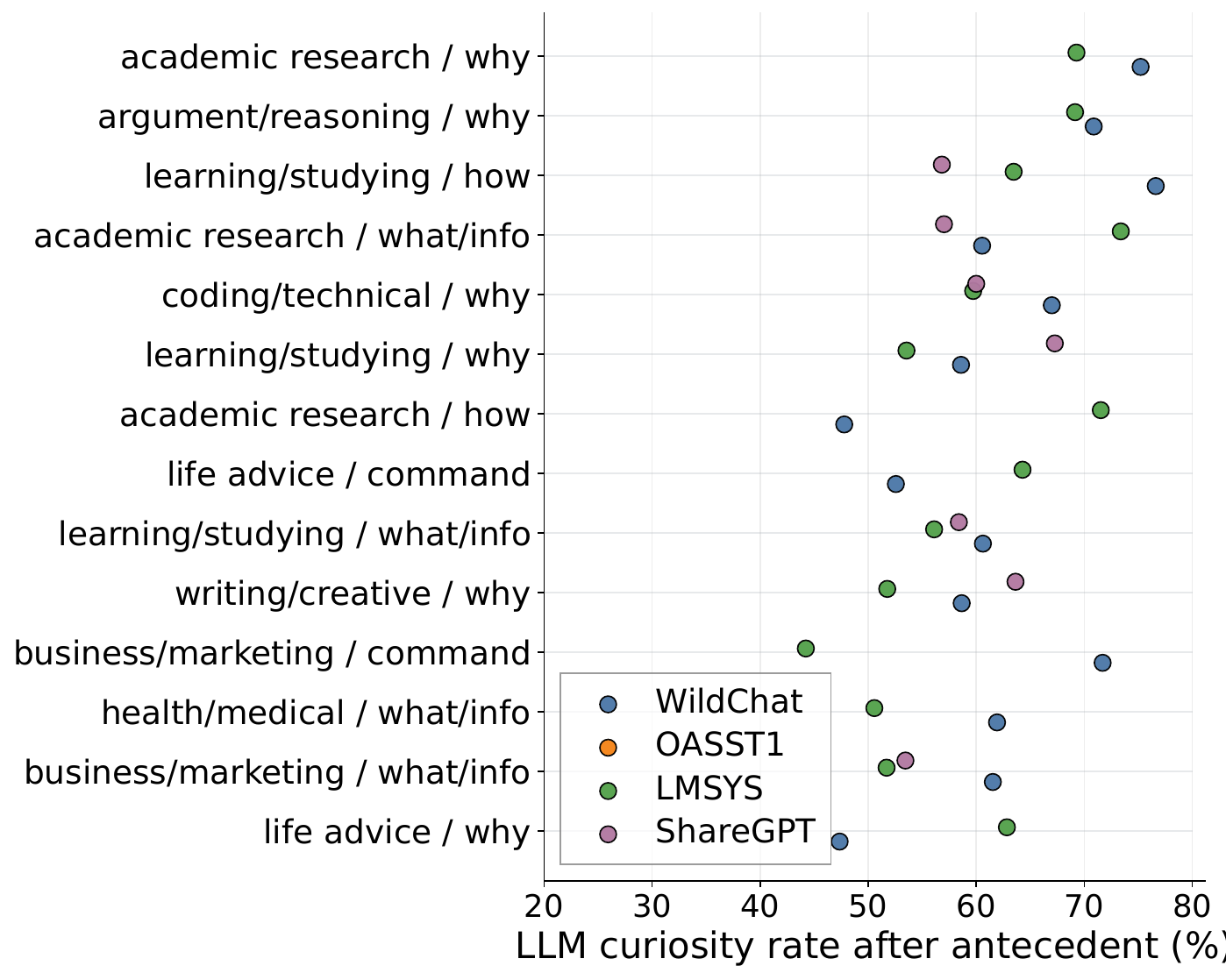}
\caption{Topic/context analysis of human questions that lead to LLM curiosity.}
\vspace{-15pt}
\label{fig:antecedent}
\end{figure}

The topic labels capture common open-ended chat contexts, such as coding, writing, academic/research, learning, life advice and so on. These contexts may invite different forms of curiosity: learning and research prompts may elicit information-gap or causal questions and advice may invite reflection or perspective-taking, and creative tasks may prompt questions about audience or intent. We assign topic and request-form labels to the preceding user turn using rule-based heuristics. Topic labels rely on keyword and phrase patterns, while request-form labels use linguistic patterns. These coarse labels support descriptive analysis rather than a validated semantic annotations. 

Fig.~\ref{fig:antecedent} shows that curiosity-oriented model questions are most common in epistemic topics, especially academic/research, argument/reasoning, and learning/studying. They often follow why/how questions, reasoning claims, research framing, and open-ended requests. They are less consistent in life-advice and business contexts, and less common in more execution-oriented settings and fully specified task requests. 

Full dataset-specific antecedent/context results and implementation are provided in Appendix~\ref{app:antecedent-analysis}.

\noindent \textbf{Model Analysis.} We also investigate model-level variation in datasets where explicit model names are available. We compute a normalized curiosity rate over real LLM questions directed to user, defined as \textbf{\(\#\text{curiosity}/(\#\text{curiosity}+\#\text{non-curiosity})\)} and exclude artifacts/unclear/problematic cases. Curiosity rates vary substantially across models: \texttt{gpt-4} and Claude models show relatively higher rates, while some LMSYS open-source models (\texttt{alpaca-13b}, \texttt{llama-2-13b-chat}, \texttt{vicuna-13b}, etc.) show lower rates. We interpret these patterns descriptively rather than as causal model rankings, since models differ in dataset coverage, user populations, prompt topics, deployment settings, and time periods. Full model-level results are in Appendix~\ref{app:model_curiosity}.

\noindent \textbf{Human Validation.} We observe substantial agreement between humans and LLM-judge on whether the LLM question is a real question (87.0\%, $\kappa=0.80$), if it is curiosity-oriented (92.0\%, $\kappa=0.82$), and its subtype (88\%). Disagreements mainly involve sensitive, leading, or confusing questions, which the LLM often labels as \textit{problematic} or \textit{unclear}, while humans more often use simpler yes/no labels. Futhermore, agreement among the annotators is also high (82.0\% for real questions and 80\% for curiosity for 20 overlapping examples). Although the shared human-human subset is small, these results suggest that both the extraction validity and curiosity labels are reasonably reliable.



\section{Downstream User Engagement and Curiosity}
We examine whether curiosity-oriented questions are associated with richer user engagement and curiosity.


\noindent \textbf{Method.} We use two complementary user signals to characterize the association between LLM curiosity and downstream user responses: \textbf{(1) User Engagement} such as follow-up question asking, exploratory questions, epistemic markers, and response length using textual proxies.  Here, epistemic markers refer to lexical cues such as why, how, I wonder, etc. These reflect epistemic engagement, which is observable reasoning in user turns, such as explanation-seeking, evidence use and so on. \textbf{(2) User Curiosity}: whether the subsequent user is curiosity-oriented using the Table~\ref{tab:curiosity-taxonomy} taxonomy, using an LLM-judge (\texttt{gpt-5-mini}). Here, the judge labels only curiosity expressed in the user's subsequent turn to model response. Note that we treat all user-side measures as observable conversational behaviors and scalable proxies, not direct measurements of users' internal curiosity states (details in Appendix~\ref{app:q2_human_proxy}). 



\noindent \textbf{Matched Analysis and Regression for User Engagement.} We use two complementary analyses to measure user engagement: (1) matched analysis and (2) adjusted regression. In both, the treatment is a curiosity-oriented model question, and the control is a non-curiosity model question. We exclude contexts where the user explicitly asked to generate questions.

For matched analysis, we pair each curiosity-oriented question with a non-curiosity question from the same model family and broad topic category. Within each block, we select the nearest control using pre-treatment features (such as prior user length, epistemic markers, exploratory questions conversation length, etc.). We then compare the next user turn across matched pairs. This creates a more comparable observational contrast, however, it does not make turns identical. Therefore, we interpret differences as adjusted associations rather than causal effects, residual confounding may remain (details in Appendix~\ref{app:matched}). 

For regression analysis, we fit adjusted models on the matched pairs. This serves as a robustness check: although matching improves comparability, observed prior differences may still remain between paired turns. For each downstream outcome, we regress the subsequent user measure on a binary curiosity indicator, controlling for the pre-treatment features used in matched analysis. The curiosity coefficient estimates the adjusted association between curiosity-oriented questioning and the subsequent user response. As with matching, we interpret these estimates as correlational rather than causal.

\noindent \textbf{LLM-Judge for User Curiosity.} We annotate matched next-user turns with an independent LLM-judge (\texttt{gpt-5-mini)}. Given the previous user turn, model question, and subsequent user turn, the judge labels whether the user expresses curiosity and its type, or whether the turn instead reflects task continuation, acknowledgment, formatting, or answering the model. Due to cost constraints, we annotate up to 2,000 matched pairs per dataset except OASST1 pairs, which in total are 468.

We additionally validate downstream user-curiosity
annotations against judgments from two human annotators, obtaining 89.9\% human-LLM agreement ($\kappa = 0.838$). This provides additional support for the LLM-based annotations, although these labels remain proxies for observable
curiosity-related behavior.

\begin{figure}
\centering
\includegraphics[width=\linewidth]{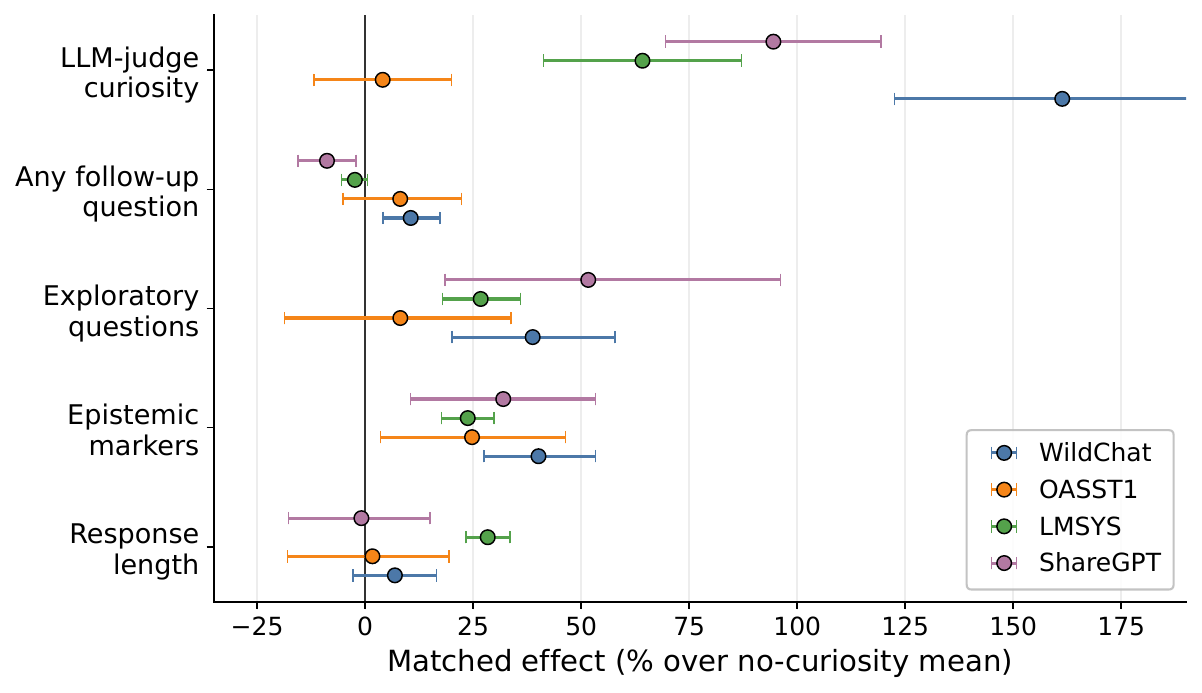}
\caption{Matched Analysis. Curiosity questions lead to more engagement and curiosity overall, however it varies by dataset. Estimates are
observational and do not establish causal effects.}
\vspace{-10.5pt}
\label{fig:dowstream1}
\end{figure}

\noindent \textbf{Results.} 
Figure~\ref{fig:dowstream1} shows that curiosity-oriented model questions are associated with stronger downstream epistemic engagement than matched non-curiosity questions. Matched effects are generally positive for LLM-judged user-curiosity as well. The strongest patterns are observed in WildChat, LMSYS, and ShareGPT and it is noisier in the smaller OASST1 dataset.

\begin{table}[t]
\centering
\small
\setlength{\tabcolsep}{2.5pt}
\begin{tabular}{@{}lrrrrr@{}}
\toprule
\textbf{Data} & \textbf{AnyQ} & \textbf{ExplQ} & \textbf{Epist.} & \textbf{Words} & \textbf{Judge} \\
\midrule
WildChat & +2.0$^{**}$ & +0.053$^{+}$ & +0.165$^{+}$ & +11.31$^{+}$ & +12.1$^{+}$ \\
OASST1   & $-$1.9        & $-$0.008          & +0.085$^{*}$   & -0.16          & +1.5          \\
LMSYS    & -3.5$^{*}$  & +0.035$^{+}$ & +0.072$^{+}$ & +5.90$^{+}$  & +9.5$^{+}$  \\
ShareGPT & -1.8        & +0.014$^{*}$   & +0.084$^{+}$ & +3.28$^{+}$  & +13.7$^{+}$ \\
\bottomrule
\end{tabular}
\caption{Regression and validation estimates for downstream user engagement. \textbf{AnyQ} is the adjusted change in next-turn question asking; \textbf{ExplQ}, \textbf{Epist.}, and \textbf{Words} are adjusted coeff. for exploratory
questions, epistemic markers, and response length; and \textbf{Judge} is the matched LLM-judge curiosity estimate.
Significance markers denote two-sided tests: $^{*}p<.05$, $^{**}p<.01$, $^{+}p<.001$.}
\label{tab:q2-regression-llm-results}
\vspace{-10pt}
\end{table}

In Table~\ref{tab:q2-regression-llm-results}, regression results show that model curiosity is more consistently associated with richer epistemic content than with simply asking another question. The AnyQ outcome is mixed, while exploratory questions, epistemic markers, response length, and LLM-judge curiosity estimates are generally positive, especially in WildChat, LMSYS, and ShareGPT. OASST1 shows weaker and noisier patterns. Overall, we observe that curiosity-oriented model questions are linked to more user engagement along with curiosity, with variation across datasets.

Above results show that associations vary across datasets. WildChat, LMSYS, and ShareGPT generally show positive
associations with downstream epistemic content, whereas OASST1 shows weaker and less consistent results. Because the datasets differ in collection
procedures, participant roles, model coverage, and selection mechanisms, we do not treat them as interchangeable populations. Moreover, matching and regression cannot eliminate unobserved factors, such as prior user interest or task complexity, that may influence both assistant questioning and
subsequent user engagement.



\subsection{Causal-Motivated Robustness Analysis for Engagement}
\label{sec:q2-causal}

This analysis moves beyond an average next-turn comparison to characterize the evidential scope of the result: whether the association remains under an alternative adjustment strategy. We treat each eligible model turn as an event: the treatment indicates whether the response at turn $t$ contains a curiosity-oriented question, with immediate outcomes measured descriptively at $t{+}1$ For each event, we compute pre-response context features.

We use inverse-probability weighting to estimate an ATT-style adjusted association~\cite{robins1992recovery, hirano2003efficient}. We model each event's probability of containing a curiosity-oriented question, $\hat e(X_i)=P(T_i=1\mid X_i)$, assign curiosity events weight $w_i=1$, and assign non-curiosity controls weight $w_i=\hat e(X_i)/(1-\hat e(X_i))$. This re-weights controls to better resemble contexts where model curiosity occurred, uses the full eligible control pool, and reduces dependence on a single matched comparison. IPW uses rule-based immediate engagement outcomes; a separate LLM judge analysis descriptively traces human-curiosity from $t{+}1$ to $t{+}5$. Because the conversations are observational, all estimates are interpreted as adjusted associations rather than causal effects; Appendix~\ref{app:q2-causal-method} details the procedure and Appendix~\ref{app:q2_covariate_balance} reports covariate-balance diagnostics.

\begin{figure}[t]
\centering
\vspace{-15pt}
\includegraphics[
    width=\linewidth,
    height=0.23\textheight,
    keepaspectratio
]{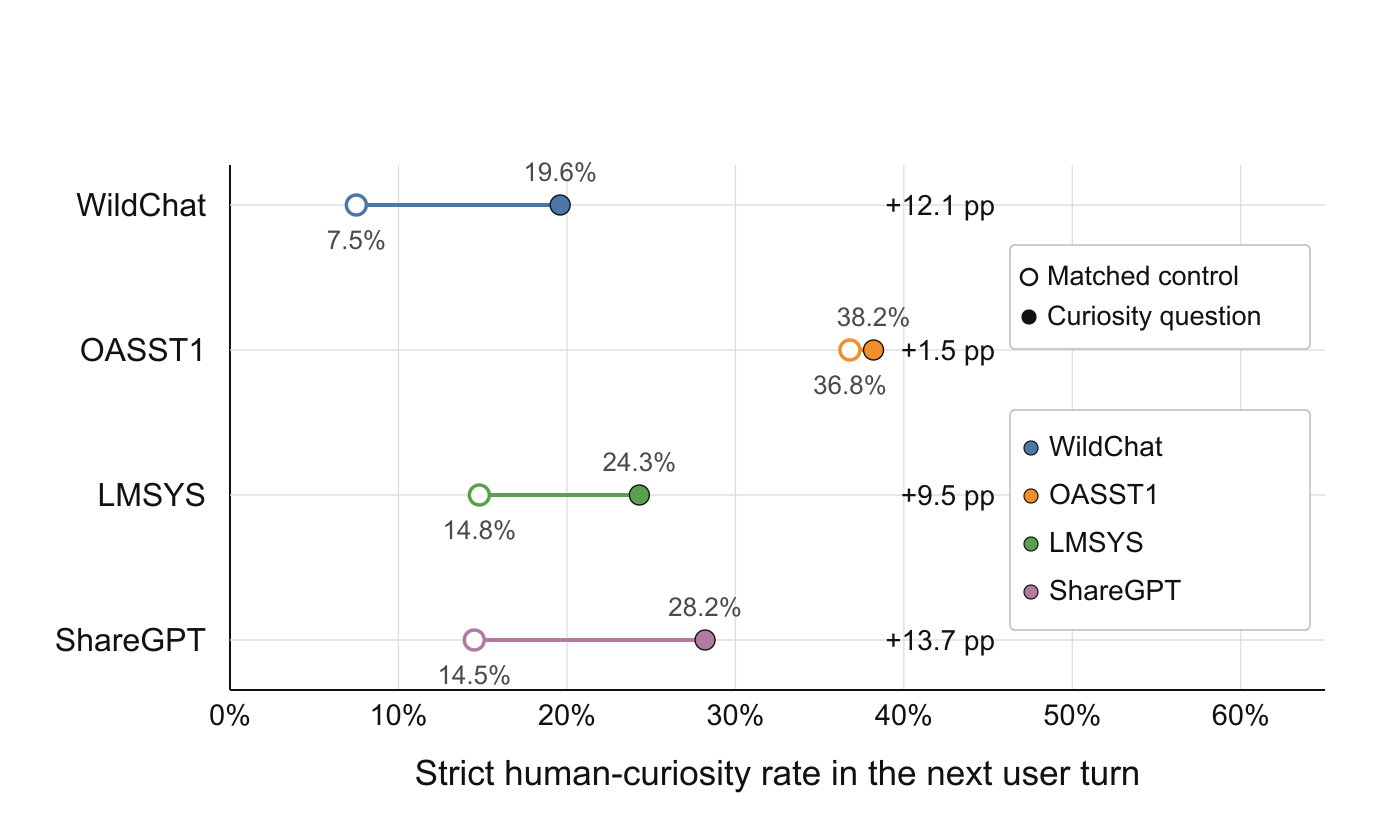}
\caption{Strict LLM-judged human-curiosity rates in matched next-user turns following curiosity-oriented model questions and clean controls.}
\label{fig:q2-causal-validation}
\vspace{-20pt}
\end{figure}

\noindent \textbf{Results.} In the pooled immediate-turn analysis, curiosity questions are followed by more user questions, exploratory questions, epistemic markers, and longer responses than clean controls, but a slightly lower probability of any follow-up question. ATT estimates provide a causal-style sensitivity analysis, but remain dataset-sensitive and should be interpreted descriptively. Matched LLM-judge validation, using a stricter human-curiosity outcome rather than marker-based epistemic engagement, finds higher next-turn human-curiosity rates (Figure~\ref{fig:q2-causal-validation}). Thus, the most consistent pattern is richer curiosity content, not simply more question-asking. We also study subtype-based changes and find that reflective-metacognitive questions show the largest increase in next-turn epistemic markers. Appendix~\ref{app:causal_subtype} shows these details. 

\noindent \textbf{Multi-Turn Persistence.} We examine judged human curiosity across five later user turns in WildChat, LMSYS, and ShareGPT. Across 1,861 valid judged turns, the treatment-control difference is negative from $t+1$ to $t+3$, approaches zero at $t+4$ ($+0.1$ pp), and becomes slightly positive at $t+5$ ($+1.4$ pp) (Figure~\ref{fig:q2-multiturn-engagement}). 
The late crossover provides an encouraging descriptive signal that model-elicited human-curiosity responses may emerge gradually. Further explanation of longitudinal analysis to study multi turn persistence and subtype-specific results appears in Appendix~\ref{app:q2-causal-extensions-q2}.

\section{Towards Useful Model Curiosity}
\label{sec: useful-curiosity-q3}
Our findings so far show that curiosity-oriented model questions can be associated with richer human engagement and curiosity. Therefore, this focuses on downstream user behavior. However, this does not mean models should always be more curious. Such behavior can be useful when they surface assumptions, invite explanation, or broaden thought processes, but harmful when they add unnecessary burden or invade privacy~\cite{horvitz1999principles, nissenbaum2004privacy, amershi2019guidelines}. Additionally, these outcomes need not coincide: a question may encourage further exploration without being preferred over a direct answer: engagement and preference may not be correlated~\cite{chen2025exploring}. Lower preference also does not always mean lower benefit: learners may favor approaches that feel more effective despite learning less~\cite{deslauriers2019measuring, kornell2008learning}. These distinctions therefore motivate a human evaluation of when curiosity-oriented questions are helpful or curiosity-awakening, and when they are burdensome or privacy-sensitive.






\begin{figure}
\centering
\includegraphics[width=0.8\linewidth]{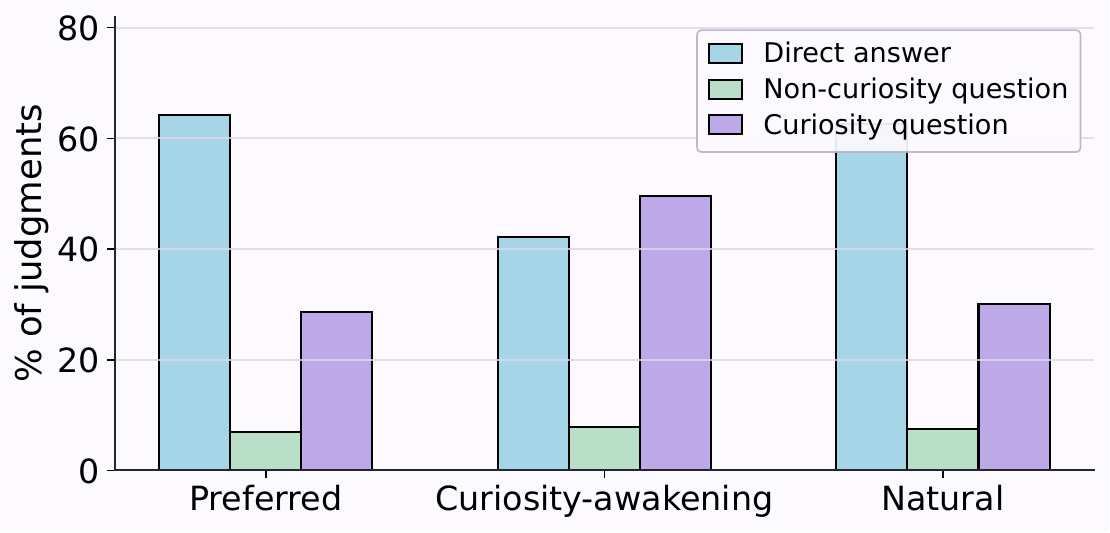}
\caption{Human Study: Participants often prefer directness, and it is low for curiosity questions. However, that differs by subtype.}
\vspace{-10.5pt}
\label{fig:human_valid}
\end{figure}

\begin{figure}
\centering
\includegraphics[width=0.8\linewidth]{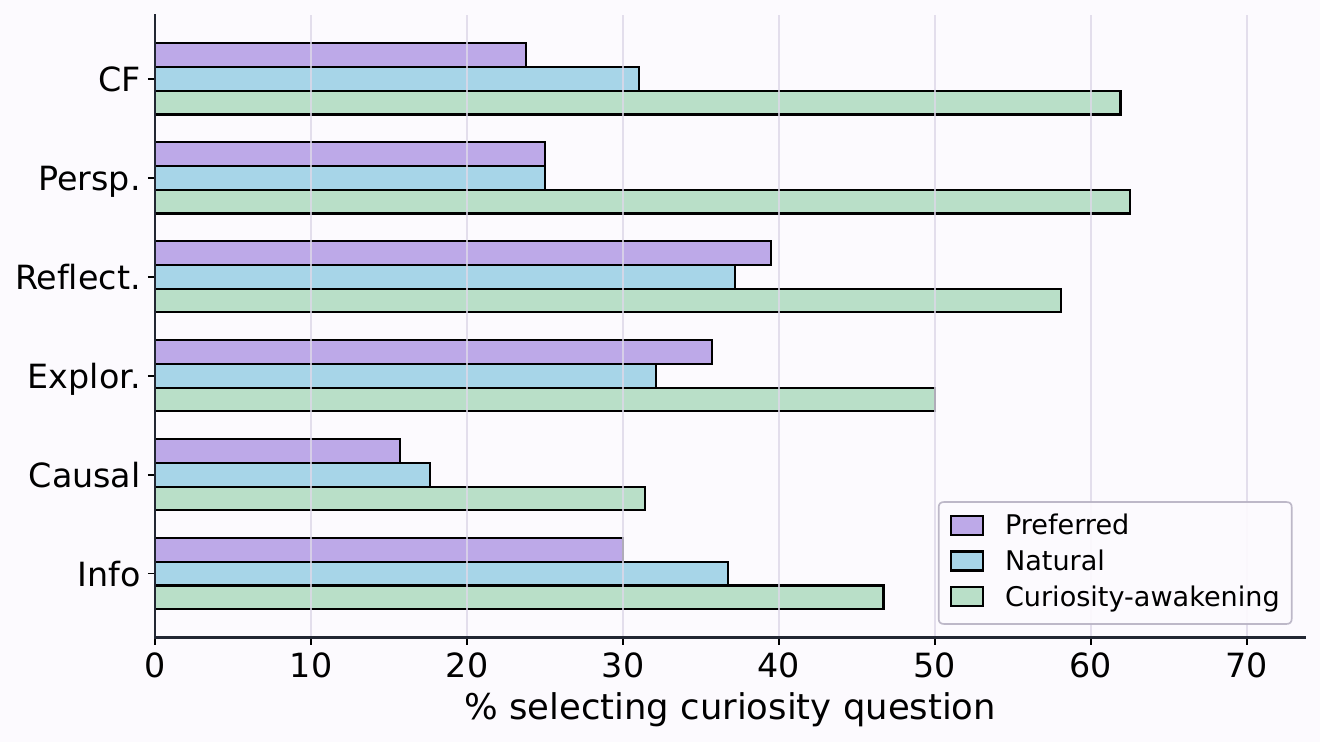}
\caption{Human-Study: Some model curiosity subtypes are preferred (reflective, exploratory) more than others (causal, counterfactual). }
\vspace{-15pt}
\label{fig:human_stacked}
\end{figure}

\subsection{Human Study}

\noindent \textbf{Study Design.} We conduct a Prolific-based\footnote{\url{https://www.prolific.com/}}  study with U.S.-based participants. Each participant reviews five/six items sampled across datasets and curiosity subtypes. Each item shows a user message with three model responses: (A) a direct answer, (B) a non-curiosity question (task clarification/engagement maintenance), and (C) a curiosity-oriented question. Participants judge: (1) preference: which response they prefer, (2) curiosity-awakening: which would make them more curious/engage further, (3) naturalness: which response feels natural, (4) annoyance/unsafe: whether the curiosity-oriented question feels annoying or unsafe and (5) follow-up needed? whether the model should/could ask a follow-up question. They also complete an attention check. We exclude low-quality responses based on failed attention checks, bot/LLM usage, or non-substantive answers. After filtering, the final analysis includes 50 participants judging 230 items (Further details are in Appendix~\ref{app:human_study_q3}). 

\noindent \textbf{Candidate Questions for Survey.} For this, we first find missed opportunities, user queries that contain uncertainty, learning goals, advice dilemmas, etc. (inspired by Fig~\ref{fig:antecedent} antecedent analysis), but models do not ask any question. However, we only take cases that are not followed by an LLM curiosity. These cases mark moments where curiosity might have helped. We also include contrast cases where a direct answer may be sufficient, allowing us to test the boundary between useful curiosity and unnecessary questioning. Using our taxonomy, we generate curiosity-oriented/non-curiosity questions using \texttt{gpt-5.5}. The direct-answer option is the original model response from the dataset.

\noindent \textbf{Results.} Fig~\ref{fig:human_valid} shows that participants most often prefer direct answers and judge them as more natural. However, curiosity-oriented questions are more often selected as making participants curious to keep thinking or asking about the topic. This suggests that model curiosity may trade off immediate directness or familiarity while increasing motivation to continue exploring. We also report item-level modal agreement among annotators in Appendix~\ref{app:human_study_agreement}; agreement is 73.5\%-77.4\% for preference, naturalness and curiosity-awakening. 

\begin{figure}
\centering
\includegraphics[width=0.8\linewidth]{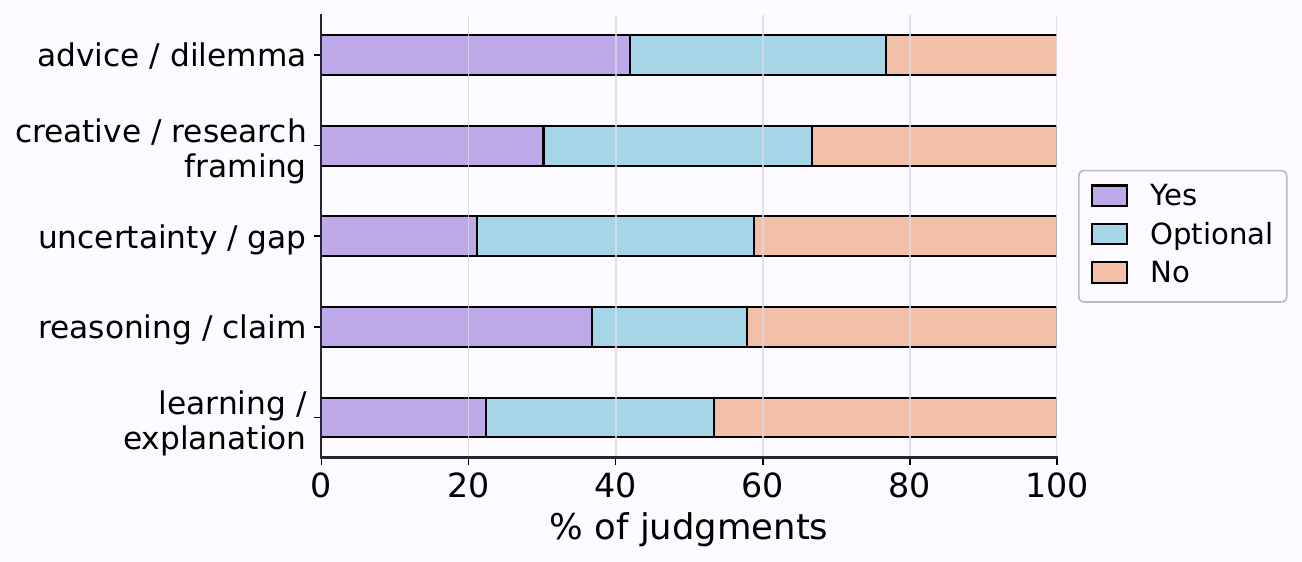}
\caption{Human-Study: Question-Asking Preferences vary by the query context. Advice/Reasoning/Creative contexts prefer more model questions.}
\vspace{-20pt}
\label{fig:question_ask_pref}
\end{figure}

The model-curiosity subtype analysis in Fig~\ref{fig:human_stacked} shows that reflective and exploratory questions are more often preferred, judged natural, and selected as curiosity-awakening. Counterfactual and perspective-taking questions also increase curiosity. However, causal has the lowest scores. Finally, Fig~\ref{fig:question_ask_pref} shows that preference for a question varies by context. Participants are most receptive to follow-up questions in queries that include advice/dilemma, creative/research-framing contexts. But, they are less receptive in explanation contexts. Overall, useful model curiosity is selective: users often prefer direct answers, however, curiosity preference can vary by context and subtype and they make people more curious and engaged.





\subsection{Modeling Useful Curiosity}
Following the human study, we test whether \textit{useful curiosity} can be predicted from observable features. We define \textit{useful curiosity} as a curiosity-oriented question that supports continued inquiry without adding burden. Since usefulness is subjective, we aggregate annotator judgments rather than treating any single judgment as ground truth. For each item $i$ and annotator $a$, we define:
\begin{equation}
u_{ia} =
\mathbb{1}\left[
C_{ia}=1 \;\land\; F_{ia}=1 \;\land\; B_{ia}=0
\right].
\end{equation}

Here, $C_{ia}$ indicates curiosity awakening, $F_{ia}$ indicates that a follow-up should or could be asked, and $B_{ia}$ indicates burden or safety concerns. So, a usefully curious question aggregates the above judgments. We exclude preference and naturalness as they capture immediate acceptability. We aggregate votes into consensus labels: \textit{validated-useful} if at least two annotators vote useful and the useful-vote rate is at least two-thirds, and \textit{validated-not-useful} for the vice-versa case. Remaining items are ambiguous. This yields 18 validated-useful, 39 validated-not-useful, and 36 ambiguous items.

\noindent \textbf{Modeling.} Using the non-ambiguous items, we frame useful curiosity as a proof-of-concept binary classification task. Each example contains a user prompt and a candidate curiosity-oriented follow-up question, labeled by human-study consensus as \textit{validated-useful} or \textit{validated-not-useful}. We the train logistic-regression classifiers and compare a majority-class baseline against three feature sets: (1) user-context affordance, capturing opportunities in the user prompt such as advice/dilemma or creative/research framing, motivated by the RQ1 antecedent analysis; (2) curiosity subtype, the taxonomy label of the question; and (3) text features, TF-IDF features from the user query and model question. Modeling details are in Appendix~\ref{app:useful_modeling}.

Table~\ref{tab:q3-useful-curiosity-modeling} shows that the majority baseline achieves high accuracy as labels are imbalanced, but has low macro-F1 and AUC. In contrast, affordance-only and TFIDF models improve macro-F1, suggesting that the user context may provide important signals for useful curiosity prediction. However, TF-IDF features are purely lexical, and accurate modeling may require context, intent, and semantic features. Because the validated set is small, we treat these results as proof of concept. But they show that useful curiosity can be framed as a supervised prediction target rather than only a qualitative intuition.


\begin{table}[t]
\centering
\small
\begin{tabular}{lrrr}
\toprule
\textbf{Model} & \textbf{Acc.} & \textbf{Macro-F1} & \textbf{AUC} \\
\midrule
Majority baseline & 0.685 & 0.406 & 0.500 \\
Affordance only & 0.652 & 0.625 & 0.663 \\
Subtype only & 0.562 & 0.559 & 0.605 \\
User + question TF-IDF & 0.668 & 0.618 & 0.670 \\
\bottomrule
\end{tabular}
\caption{Proof-of-concept prediction of item-level validated useful curiosity. Accuracy is affected by class imbalance, macro-F1 better reflects recovery of useful-curiosity cases.}
\label{tab:q3-useful-curiosity-modeling}
\vspace{-20pt}
\end{table}



\section{Lessons Learned and Actionable Takeaways} 


Through our experiments and analyses, we learned several valuable insights:

\noindent \textbf{Curiosity is not the same as asking questions.} Many model questions serve clarification or engagement maintenance. However, evaluating curiosity requires theory-grounded distinctions that separate them from generic questions.

\noindent \textbf{Model curiosity exists in real human-LLM interactions, but is not the default.} LLMs do ask curiosity-oriented questions, but these remain rare. When curiosity appears, it is concentrated in information-gap, causal-explanatory, and exploratory-broadening forms, while reflective and perspective-taking are much rarer but important. Therefore, future systems should work towards explicitly supporting more diverse forms of curiosity.


\noindent \textbf{Model curiosity is associated with richer user epistemic engagement and curiosity.} Users respond with more curious and epistemic content, especially after reflective and exploratory model questions. Future work should test these patterns in controlled experiments to establish causal effects on downstream user behavior.

\noindent \textbf{Useful curiosity requires selective questioning.} Users often prefer direct answers, but targeted reflective and exploratory questions can help them continue thinking, especially in advice/dilemma, creative, and research-framing contexts. Future systems should model useful curiosity as timely, relevant, low-burden questioning that supports continued inquiry without interrupting the user's goal. A preliminary prompting pilot further suggests
that curiosity-oriented questioning can be elicited
through explicit instructions (Appendix~\ref{app:prompting_pilot}).
However, increasing the frequency of such questions
does not necessarily improve their usefulness.


\section{Conclusion}
We study curiosity in LLM during human-LLM conversations, grounding it in prior theories of information gaps, explanation-seeking, reflection, and exploration. We first introduce a taxonomy that distinguishes curiosity-oriented questions from other conversational artifacts. Across four public datasets, we find that curious LLM questions are associated with richer downstream engagement and curiosity, although the strength and consistency of these associations
vary across datasets. We further define \textit{useful curiosity} and show through a human study that reflective and exploratory model questions can encourage continued curiosity. Together, these findings suggest that future LLMs should not simply ask more questions, but learn when and how to ask questions that support better human-LLM collaboration.

\section*{Limitations}
\noindent\textbf{Dataset and Multilingual Coverage.}
Our datasets span different platforms, collection periods, model families, conversation structures, and user populations. While they include some multilingual conversations, our taxonomy and evaluation are developed primarily for English-language behavior, reflecting author expertise and the practical constraints of human validation. Because curiosity may be expressed differently across languages and cultures~\citep{borah-etal-2025-curious}, the reported categories and rates may not transfer uniformly across communities. This is also related to broader WEIRD biases in NLP and AI evaluation~\citep{mihalcea2025ai}, we therefore view multilingual and cross-cultural validation of model curiosity as an important direction for future work.

\noindent\textbf{Coverage of Newer Models.}
Our results characterize the models represented in the available datasets and may not capture recent changes in newer  models. Evaluating these models will be interesting and require further interaction data. We view this as an important direction for future work.

\noindent\textbf{LLM-Judge Validation.}
The LLM judge enables contextual evaluation at scale, but its labels may reflect model-specific biases or conflate curiosity with related behaviors such as uncertainty, elaboration, or task continuation. Human validation is helpful; however, it is on a limited range of annotator backgrounds. 

\noindent\textbf{Scope of the Findings.}
The study measures curiosity as it is expressed through conversation, not as an internal psychological state. Moreover, naturally occurring conversation data support associations rather than causal conclusions. Although the human study is helpful, it is limited in terms of size and diversity. Controlled studies with diverse participants and languages will establish when model curiosity reliably benefits users of all communities~\cite{karamolegkou2026nlp}.

\section*{Ethical Considerations}

Curiosity-oriented questions can support reflection, but they may also pressure users to disclose sensitive information or feel intrusive, burdensome, or manipulative, especially in contexts involving health, identity, relationships, or personal uncertainty. Our human study therefore evaluates whether questions are helpful, unnecessary, annoying, invasive, or unsafe, rather than measuring engagement alone. Participants provided consent, and their judgments are reported only in aggregate. Because public conversation logs may still contain self-disclosed personal information despite dataset-level redaction, we do not attempt to identify users or infer private attributes, and we paraphrase examples where appropriate. Our work should not be used to optimize models simply for longer conversations or greater engagement. Systems that repeatedly probe users, exploit uncertainty, or infer sensitive traits could undermine privacy and user autonomy. Curiosity-oriented questions should remain relevant, low-burden, and responsive to the user's goal, with direct answers provided when further questioning is unnecessary.

Our analysis may also reflect biases in the underlying datasets, models, and annotation process. Model-generated questions may reproduce social, cultural, or linguistic biases learned during training or alignment~\cite{borah2024towards}. Human judgments of what counts as useful, natural, intrusive, or burdensome may also vary across demographic and cultural groups. Furthermore, LLM-LLM  and Human-LLM interactions also lead to harmful behaviors such as polarization or echo chambers~\cite{borah-etal-2026-persuasion}, which maybe displayed in the corpora we use. We therefore interpret our taxonomy and usefulness results as a step toward measuring model curiosity, not as a universal standard for how systems should question users across all communities. Future work should evaluate model curiosity across broader languages, cultures, accessibility needs, and high-stakes contexts before deploying curiosity-oriented behaviors widely.

\section*{Acknowledgments}
We thank the anonymous reviewers for their constructive feedback. We are also grateful to the members of the Language and Information Technologies Lab at the University of Michigan for their valuable input and insightful discussions during the early stages of the project. This project was partially funded by a grant from OpenAI, an award from the Robert Wood Johnson Foundation (\#80345), and a grant from the Survival and Flourishing Fund. Any opinions, findings, and conclusions or recommendations expressed in this material are those of the authors and do not necessarily reflect the views of OpenAI or the Robert Wood Johnson Foundation or the Survival and Flourishing Fund.

\bibliography{main}

\appendix


\section{Taxonomy}
\label{taxonomy}

\subsection{Theoretical grounding}
\label{app:theory}

Table~\ref{tab:curiosity-taxonomy-full} presents the full taxonomy, including each question type, subtype, and its theoretical grounding.

\subsection{Taxonomy Construction Extended}
\label{app:taxonomy_extended}

Building on this distinction, we construct a taxonomy that separates curiosity-oriented questions from other question functions in human-LM conversations. We ground our taxonomy in theory-driven, drawing on information-gap theories of curiosity~\cite{loewenstein1994psychology}, epistemic curiosity~\cite{litman2005curiosity}, self-explanation in learning~\cite{chi1994eliciting}, and mixed-initiative tutoring dialogue~\cite{graesser2005autotutor}. We distinguish six curiosity-oriented subtypes based on the epistemic move made by the question. \textit{Information-gap} questions ask for missing knowledge or context that could improve understanding. \textit{Causal-explanatory} questions go beyond missing facts to ask why or how something happens, focusing on reasons, mechanisms, or explanations. \textit{Exploratory-broadening} questions widen the space of inquiry by inviting additional interpretations, implications, examples, or possible directions. \textit{Reflective-metacognitive} questions turn attention back to the user's own reasoning process, asking them to examine assumptions, evidence, confidence, uncertainty, or goals. \textit{Perspective-taking} questions shift the frame outward by inviting the user to consider another person, stakeholder, audience, or viewpoint. Finally, \textit{counterfactual-hypothetical} questions ask the user to reason under imagined or altered conditions, making explicit how a situation might change if key assumptions or circumstances were different.


We also include non-curiosity categories that serve different interactional functions. These are drawn from previous work on clarification in conversational systems~\cite{rahmani2023survey} and task-oriented information-seeking dialogue~\cite{feng2023towards}. They also include \textit{context requests} for missing task information, \textit{clarification repair} for resolving ambiguity, \textit{task management} for questions on format, tone, language, or tool use, and \textit{engagement maintenance} for generic continuation prompts. We also separate potentially problematic questions, including \textit{privacy probing} and \textit{leading or steering}, and generated artifacts such as quiz items, roleplay dialogue, interview templates, or malformed fragments that contain question marks but are not LLM-to-user questions. These markers are important for large-scale interaction datasets such as WildChat~\cite{zhao2024wildchat}.

\subsection{Operationalization of Taxonomy and LLM Judge Details}
\label{app:llmjudge}

\noindent \textbf{Rule-based preprocessing and extraction.} Generated artifacts are very common across datasets. In WildChat, many apparent questions occur inside generated writing, scripts, roleplay, quizzes, interview prompts, or templates, rather than being addressed to the user. Without this filtering, counts of model questioning would be inflated by text that merely contains questions but does not function as a conversational question.

 We segment each assistant response into sentences and retain spans that contain question marks or interrogative forms such as \textit{what}, \textit{why}, \textit{how}, \textit{which}, \textit{when}, \textit{where}, \textit{who}, \textit{do you}, \textit{would you}, or \textit{could you}. We also remove obvious artifacts, including malformed extractions, questions copied from the user turn, questions inside code blocks or quoted text, generated example questions, quiz/list items, and questions embedded in fictional dialogue rather than directed to the user. For each retained candidate, we store the extracted question together with the previous user turn  and the full assistant response for contextual judging.

\noindent \textbf{LLM judge prompt.} We then use an LLM judge to classify each candidate question in context. The judge receives the previous user turn, the full model response, and the extracted question. It determines whether the extracted question is a real LLM-to-user question, assigns a top-level label (\textit{curiosity}, \textit{not curiosity}, or \textit{problematic}), and, for curiosity-oriented questions, assigns one of the six curiosity subtypes. The prompt instructs the judge to distinguish substantive curiosity from task clarification, formatting questions, shallow engagement, rhetorical questions, and generated or quoted question artifacts.

\textit{Judge input and output:}
Each classification instance contains the preceding user
turn, the full assistant response, and the extracted
candidate question. The judge follows the taxonomy and
boundary-case instructions to determine whether the
question is directed to the user, assign a top-level
status, and select the best-fitting subtype. Generated
artifacts are excluded from the retained question set.

\begin{figure*}[t]
\centering
\begin{tcolorbox}[
  width=\textwidth,
  colback=red!5,
  colframe=red!60!black,
  title={LLM Judge Prompt},
  fonttitle=\bfseries,
  arc=2mm,
  boxrule=0.8pt,
  left=1.5mm,
  right=1.5mm,
  top=1mm,
  bottom=1mm
]
\scriptsize
\setlength{\parskip}{2pt}

\textbf{System prompt.} You are an expert annotator for human--LLM conversation research. Classify assistant questions and return one judgment for every item id.

First decide whether the extracted question is truly the assistant asking the user a live question. Exclude generated content, roleplay/dialogue, quiz or game mechanics, examples/templates, generated writing, embedded questions, and malformed fragments.

\textbf{Top-level status.}
\texttt{curiosity}: opens, deepens, examines, or resolves an epistemic gap.
\texttt{not\_curiosity}: task or conversation management without epistemic inquiry.
\texttt{problematic}: probes sensitive information or steers the user.
\texttt{unclear}: insufficient context or ambiguous function.

  \textbf{Question types:}
  \texttt{causal\_explanatory} asks why/how something happens or works
  (e.g., ``Why do you think this result occurred?'');

  \texttt{counterfactual\_hypothetical} asks about an imagined changed condition
  (e.g., ``How would this change if the audience were beginners?'');

  \texttt{exploratory\_broadening} invites alternatives, implications, or broader directions
  (e.g., ``What other approaches might be worth considering?'');

  \texttt{perspective\_taking} asks about another person, group, stakeholder, or viewpoint
  (e.g., ``How might the other person interpret this?'');

  \texttt{reflective\_metacognitive} asks about the user's assumptions, goals, evidence, uncertainty, criteria,
  or reasoning
  (e.g., ``What tradeoff matters most to you here?'');

  \texttt{information\_request} asks for missing facts or context needed to continue
  (e.g., ``What topic should this be about?'');

  \texttt{clarification\_repair} asks the user to clarify ambiguous wording or an underspecified request
  (e.g., ``What do you mean by `better' here?'');

  \texttt{task\_management} asks about output format, style, length, audience, workflow, or constraints
  (e.g., ``Do you want this as bullets or a table?'');

  \texttt{engagement\_maintenance} is a generic continuation or helpfulness prompt
  (e.g., ``Would you like me to continue?'');

  \texttt{privacy\_probing} asks for sensitive personal information
  (e.g., ``What is your exact address or medical history?'');

  \texttt{leading\_steering} pushes the user toward a specific answer, belief, action, or disclosure
  (e.g., ``Don't you agree this is clearly the best choice?'');

\textbf{Rules.}
A question mark alone is not enough. Generated questions, fictional dialogue, quiz items, templates, quoted text, and malformed fragments are not live assistant-to-user questions. If \texttt{keep} is false, \texttt{artifact\_type} must not be \texttt{none}, \texttt{status} must be \texttt{unclear}, and \texttt{question\_type} must be \texttt{unknown\_question}. If \texttt{keep} is true, \texttt{artifact\_type} must be \texttt{none}.


\end{tcolorbox}
\caption{Prompt used for LLM-judge annotation of human-LLM conversations.}
\label{fig:llm-judge-prompt}
\end{figure*}

\paragraph{Human validation.} Finally, we validate the judge on a smaller annotated subset consisting of 260 unique candidate questions and 300 total annotations. Three human annotators each label 100 examples, with 20 shared examples included across all annotators for inter-annotator agreement. Annotators see the previous user turn, the assistant response, and the extracted question. They label whether the extracted question is real, whether it is curiosity-oriented, and, when applicable, which curiosity subtype best captures its dominant function. We use this subset to estimate human--LLM agreement, measure annotator agreement on the shared items, and check whether the taxonomy is applied consistently. Fig~\ref{fig:human-annotation-instructions} contains the instructions used for human validation of LLM-judge instructions.

\begin{figure}[t]
\centering
\begin{tcolorbox}[
    colback=blue!5,
    colframe=blue!70!black,
    title=\textbf{Human Annotation Instructions},
    fonttitle=\bfseries,
    arc=2mm,
    boxrule=0.8pt
]
\scriptsize
Annotators were shown a short conversation context, including the user query, the assistant turn, and the extracted assistant question. They answered the following questions for each item:

\begin{enumerate}[leftmargin=*]
    \item \textbf{Real assistant question:} Is the extracted text a real assistant-to-user question in the conversation? Mark \texttt{yes}, \texttt{no}, or \texttt{unclear}. Mark \texttt{no} if the question is part of generated content, quoted text, a template, roleplay dialogue, a quiz/game, code, or another artifact.

    \item \textbf{Curiosity-oriented question:} If the extracted text is a real assistant question, does it invite the user to seek, explain, reflect on, compare, question, or explore knowledge, assumptions, evidence, perspectives, alternatives, or implications? Mark \texttt{yes}, \texttt{no}, or \texttt{unclear}. Mark \texttt{no} for task clarification, formatting requests, generic continuation prompts, or workflow management.

    \item \textbf{Curiosity subtype:} If the question is curiosity-oriented, assign the best-fitting subtype: \texttt{information\_gap}, \texttt{causal\_explanatory}, \texttt{exploratory\_broadening}, \texttt{counterfactual\_hypothetical}, \texttt{perspective\_taking}, or \texttt{reflective\_metacognitive}. Leave blank or write \texttt{none} otherwise.

    \item \textbf{Notes:} Optionally mention anything confusing, ambiguous, artifact-like, sensitive, leading, or difficult to classify.
\end{enumerate}
\end{tcolorbox}
\caption{Human annotation instructions for validating assistant-question extraction and curiosity labels.}
\label{fig:human-annotation-instructions}
\end{figure}



\subsection{Boundary Cases Between Curiosity Subtypes and/or Generic Questions}
\label{app:boundary_case}


  \paragraph{Information-gap vs. task clarification.}
  Information-gap curiosity refers to questions that open or pursue a substantive missing piece in the user's
  topic, reasoning, or decision context, rather than minimal clarification needed to execute the task. For
  example, asking what evidence, assumptions, goals, or constraints would change the interpretation is treated as
  information-gap curiosity, whereas asking for a missing file, format, length, or required input is treated as
  task clarification.

  \paragraph{Exploratory-broadening vs. counterfactual-hypothetical.}
  Exploratory-broadening questions expand the space of possible options, interpretations, implications, or next
  directions without specifying a changed condition. Counterfactual-hypothetical questions explicitly introduce
  an imagined change to the situation and ask what would follow. For example, ``What other approaches might be
  worth considering?'' is exploratory-broadening, whereas ``How would your answer change if the audience were
  beginners?'' is counterfactual-hypothetical.

  \paragraph{Information-gap vs. causal-explanatory.}
  Information-gap questions ask what missing information would help understand or decide something. Causal-
  explanatory questions specifically ask why something happened, how it works, or what mechanism explains it. For
  example, ``What information would help you decide?'' is information-gap curiosity, whereas ``What factors might
  explain this outcome?'' is causal-explanatory.

  \paragraph{Exploratory-broadening vs. perspective-taking.}
  Exploratory-broadening questions invite additional options or directions in general. Perspective-taking
  questions specifically ask the user to consider another person, group, audience, stakeholder, or viewpoint. For
  example, ``What other interpretations are possible?'' is exploratory-broadening, whereas ``How might someone
  with the opposite view respond?'' is perspective-taking.

  \paragraph{Reflective-metacognitive vs. information-gap.}
  Reflective-metacognitive questions ask users to inspect their own goals, assumptions, uncertainty, criteria, or
  reasoning process. Information-gap questions ask for missing information in the topic or situation. For
  example, ``What assumption are you making here?'' is reflective-metacognitive, whereas ``What information is
  missing from the problem statement?'' is information-gap curiosity.

  \paragraph{Reflective-metacognitive vs. perspective-taking.}
  Reflective-metacognitive questions focus on the user's own reasoning or decision process. Perspective-taking
  questions focus on another viewpoint or stakeholder. For example, ``What would make you feel confident in this
  decision?'' is reflective-metacognitive, whereas ``How might your audience interpret this choice?'' is
  perspective-taking.

\begin{table*}[!ht]
\centering
\footnotesize
\setlength{\tabcolsep}{3.5pt}
\renewcommand{\arraystretch}{1.15}
\begin{tabular}{@{}p{0.20\textwidth}p{0.23\textwidth}p{0.34\textwidth}p{0.17\textwidth}@{}}
\toprule
\textbf{Question type} & \textbf{Theoretical grounding} & \textbf{Operational definition} & \textbf{Example} \\
\midrule

Curiosity: Information-gap
& Information-gap theory; epistemic curiosity~\citep{loewenstein1994psychology,litman2005curiosity}
& Asks about missing knowledge that could deepen understanding or surface an epistemic gap, rather than merely filling a task slot.
& ``What problem does it solve?'' \\

Curiosity: Causal-explanatory
& Explanation-seeking; explanations as understanding~\citep{chi1994eliciting,lombrozo2006structure}
& Asks for causes, mechanisms, reasons, or why/how something happens.
& ``Why did this happen?'' \\

Curiosity: Exploratory broadening
& Epistemic exploration; diversive and specific curiosity~\citep{berlyne1966curiosity,litman2010validity}
& Broadens inquiry toward alternatives, implications, extensions, or new angles.
& ``What else might explain this?'' \\

Curiosity: Reflective-metacognitive
& Metacognition; epistemic reflection; uncertainty monitoring~\citep{flavell1979metacognition,nelson1990metamemory}
& Prompts reflection on assumptions, evidence, confidence, uncertainty, values, or reasoning.
& ``What assumptions are you making?'' \\

Curiosity: Perspective-taking
& Perspective-taking; role-taking theory; empathy~\citep{davis1983measuring,galinsky2000perspective}
& Invites the user to consider another person, stakeholder, viewpoint, intention, or interpretive frame.
& ``How might they interpret this?'' \\

Curiosity: Counterfactual-hypothetical
& Counterfactual reasoning; hypothetical exploration; epistemic simulation~\citep{kahneman1986norm,roese1997counterfactual}
& Asks the user to reason about changed conditions, alternative possibilities, or what would happen if something were different.
& ``What if the context changed?'' \\

\midrule

Non-curiosity: Context request
& Grounding in communication; task-oriented information gathering~\citep{clark1991grounding,aliannejadi2019asking}
& Asks for missing task information needed to produce an answer; primarily instrumental rather than exploratory.
& ``What topic should this be about?'' \\

Non-curiosity: Clarification repair
& Conversation repair; clarification questions~\citep{schegloff1977preference,rahmani2023survey}
& Asks because the user's prior turn is ambiguous, underspecified, or difficult to interpret.
& ``What do you mean by X?'' \\

Non-curiosity: Task management
& Task-oriented dialogue; conversational grounding~\citep{young2013pomdp,wen2017network}
& Asks about output constraints such as format, length, tone, style, language, tool choice, or programming language.
& ``What format should I use?'' \\

Non-curiosity: Engagement maintenance
& Engagement management; relational agents~\citep{sidner2005explorations,bickmore2005establishing}
& Keeps the interaction open without deepening inquiry or inviting substantive reflection.
& ``Would you like me to continue?'' \\

\midrule

Problematic: Privacy probing
& Privacy-sensitive information seeking; safety risks~\citep{nissenbaum2004privacy}
& Requests sensitive personal information in a potentially intrusive or unsafe way.
& ``Can you share personal details?'' \\

Problematic: Leading or steering
& Leading questions; biased or steering interaction~\citep{loftus1974reconstruction}
& Pushes the user toward a particular answer rather than opening inquiry.
& ``Isn't this clearly best?'' \\

Unclear: Unknown
& Residual annotation category
& Appears addressed to the user, but its interactional function cannot be classified with sufficient confidence.
& ``Could it be?'' \\

Preprocessing: Generated artifact
& Dataset preprocessing; artifact filtering
& Contains a question mark but is not a real assistant-to-user question, such as quiz items, templates, roleplay dialogue, generated writing, or malformed fragments.
& ``What is a black hole?'' \\

\bottomrule
\end{tabular}
\caption{Taxonomy of assistant-to-user questions. Curiosity-oriented questions are grounded in theories of information gaps, explanation-seeking, epistemic exploration, metacognition, perspective-taking, and counterfactual reasoning. Non-curiosity questions capture task, clarification, engagement, and other interactional functions. Generated artifacts are excluded before classification.}
\label{tab:curiosity-taxonomy-full}
\end{table*}

\section{Dataset Details and Preprocessing}
\label{app:data}

We analyze four public conversation datasets: WildChat~\citep{zhao2024wildchat}, OASST1~\citep{kopf2023openassistant}, LMSYS-Chat-1M~\citep{zheng2024lmsys}, and ShareGPT\footnote{\url{https://sharegpt.com/}}. Together, they cover diverse human-LLM interactions, including open-ended assistance, technical help, writing, learning, advice-seeking, and multi-turn collaboration. Our unit of analysis is not every conversation turn, but extracted assistant-question events. We first identify candidate questions from model turns, then use the LLM judge to remove artifacts and classify retained questions.

 \begin{table*}[ht]
  \centering
  \small
  \setlength{\tabcolsep}{4pt}
  \begin{tabular}{lrrrr}
  \toprule
  \textbf{Dataset} & \textbf{Candidate Qs} & \textbf{Kept real Qs} & \textbf{Dropped/artifact} &
  \textbf{Curiosity Qs} \\
  \midrule
  WildChat & 75,543 & 50,989 (67.5\%) & 24,554 (32.5\%) & 19,755 (26.2\%) \\
  OASST1 & 3,704 & 3,116 (84.1\%) & 588 (15.9\%) & 876 (23.7\%) \\
  LMSYS & 202,552 & 159,918 (79.0\%) & 42,634 (21.0\%) & 33,938 (16.8\%) \\
  ShareGPT & 28,110 & 21,462 (76.4\%) & 6,648 (23.6\%) & 6,190 (22.0\%) \\
  \bottomrule
  \end{tabular}
  \caption{Question-event preprocessing statistics. Candidate Qs are assistant-question spans extracted by rule-
  based preprocessing. Kept real Qs are retained by the LLM judge as real assistant-to-user questions. Curiosity
  Qs are retained questions classified as curiosity-oriented; percentages are relative to candidate questions.}
  \label{tab:dataset_preprocessing}
  \end{table*}

WildChat contributes a large set of open-ended user--assistant conversations and contains the highest proportion of curiosity-oriented candidate questions after extraction. OASST1 is smaller but has relatively high retention after artifact filtering, reflecting fewer generated-question artifacts in the extracted set. LMSYS contributes the largest number of candidate question events and therefore dominates the absolute count of retained real questions, although its curiosity rate is lower. ShareGPT contains many writing, roleplay, and generated-content interactions, so artifact filtering is important for separating live assistant-to-user questions from questions embedded inside generated text. Table~\ref{tab:dataset_preprocessing} illustrates dataset details, including, candidate qs, kept real qs, dropped artifact and curiosity qs.

Unlike the other three datasets, OASST1 is a human-authored conversation corpus in which contributors assume user and assistant roles. Although its assistant responses are not generated by LLMs, the dataset provides a useful comparison for studying assistant-style question-asking behavior. We therefore report its results separately and analyze its downstream engagement patterns cautiously, given differences in data collection, authorship, and interaction structure.

\begin{figure}[t]
\centering
\begin{tcolorbox}[
colback=blue!5,
colframe=blue!70!black,
title={Dataset Examples After Question Extraction},
fonttitle=\bfseries,
arc=2mm,
boxrule=0.8pt
]
\scriptsize
Each example shows the local context used by the judge: the previous user turn, the extracted assistant question, and the resulting label.

\vspace{3pt}
\textbf{WildChat.}
User: ``Can you help me think through whether this argument is fair?''
Assistant question: ``What assumptions are you making about the other side's position?''
Label: \texttt{curiosity} / \texttt{reflective\_metacognitive}. 

\vspace{4pt}
\textbf{OASST1.}
User: ``Can you simplify your answer about consciousness and redness?''
Assistant question: ``How, then, did redness come to exist?''
Label: \texttt{curiosity} / \texttt{causal\_explanatory}. 

\vspace{4pt}
\textbf{LMSYS.}
User: ``This code is not correct.''
Assistant question: ``Could you please provide more information about why it is not working correctly?''
Label: \texttt{not\_curiosity} / \texttt{clarification\_repair}. 

\vspace{4pt}
\textbf{ShareGPT.}
User: ``Continue the conversation between two characters.''
Assistant question: ``What should we do now?''
Label: \texttt{unclear} / \texttt{unknown\_question}, artifact type \texttt{roleplay\_or\_dialogue}; the question belongs to generated dialogue rather than the live assistant--user interaction.
\end{tcolorbox}
\caption{Examples of extracted assistant questions and their assigned labels across datasets.}
\label{fig:dataset-extraction-examples}
\end{figure}

\section{Q1 - Results}

\subsection{Topic and Antecedent Analysis}

\noindent \textbf{Topic and request-form annotation.}
We characterize the context preceding each assistant question along two dimensions: broad topic and request form. Topic categories are assigned through keyword and phrase matching, while request forms are identified through linguistic patterns in the preceding user turn. We use these heuristic labels to compare the contexts associated with curiosity-oriented questioning across datasets. They are not intended as exhaustive or mutually exclusive representations of user intent.

\label{app:antecedent-analysis}
\noindent \textbf{Results.} Across datasets, curiosity-oriented questions are not distributed randomly across user contexts. They are more likely to follow user turns that contain epistemic or interpretive openings, such as reasoning claims, academic or research framing, learning goals, advice dilemmas, health or business questions, and underspecified generative requests.

In WildChat, the highest curiosity rates occur after \textit{argument/reasoning} turns (69.5\%), followed by \textit{academic/research} (60.7\%), \textit{health/medical} (58.7\%), and \textit{learning/studying} (58.2\%). Similar patterns appear across other datasets. In LMSYS, \textit{academic/research} is the strongest antecedent context (64.5\%), while health, business, life advice, learning, and reasoning contexts also show elevated curiosity rates. In ShareGPT, \textit{argument/reasoning} again has the highest rate (58.2\%), followed by health, life advice, academic/research, and business/marketing. OASST1 is smaller, but shows the same broad trend, with high rates for argument/reasoning, writing/creative, and learning/studying contexts.

\begin{figure}
\centering
\includegraphics[width=\linewidth]{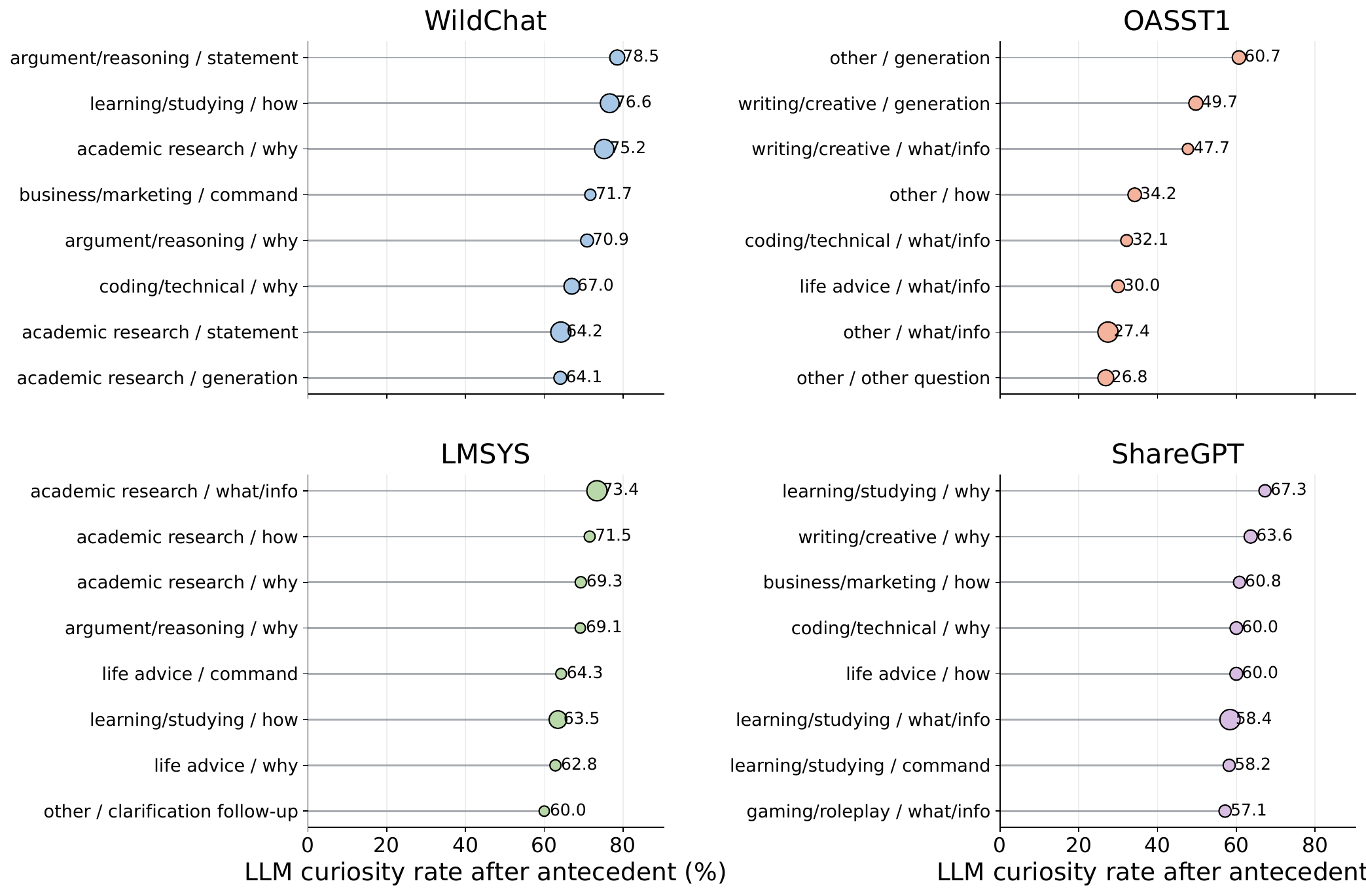}
\caption{Topics and contexts that lead to LLM curiosity across datasets. }
\vspace{-10.5pt}
\label{fig:topic_context_antecedent}
\end{figure}

User request form also matters. Across datasets, explicit \textit{why} and \textit{how} questions are among the strongest antecedents of curiosity-oriented assistant questions. For example, in WildChat, both \textit{why-causal} and \textit{how-procedural} turns lead to curiosity-oriented questions in about 57\% of cases. In LMSYS and ShareGPT, \textit{why-causal} turns are also among the strongest request-form signals. Open-ended generation requests can also elicit curiosity, especially when the user leaves goals, audience, constraints, or intended effect underspecified.

The strongest topic-by-request-form cells further support this interpretation. Across datasets, high-curiosity cells include combinations such as \textit{academic/research + why-causal}, \textit{learning/studying + how-procedural}, \textit{argument/reasoning + why-causal}, and \textit{writing/creative + request-generation}. These are contexts where the user is not merely asking for task completion, but opening space for explanation, interpretation, framing, or problem refinement.

Overall, this analysis suggests that current LLMs are more likely to ask curiosity-oriented questions when the previous user turn contains an epistemic, interpretive, or underspecified-task opening. At the same time, the dominant curiosity subtype following these contexts is usually \textit{information-gap curiosity}, suggesting that models most often respond by asking for missing information rather than by prompting deeper reflection, perspective-taking, or counterfactual reasoning.

\noindent \textbf{Model analysis.} We also investigate model-level variation in datasets that provide explicit assistant model names. For each model, we compute a normalized curiosity rate over real, classified LLM questions, defined as \textbf{\(\#\text{curiosity}/(\#\text{curiosity}+\#\text{non-curiosity})\)}. We exclude artifacts, unclear cases, and problematic cases from the above computation. We observe substantial variation across models: \texttt{gpt-4} has a higher curiosity rate than \texttt{gpt-3.5-turbo}, Claude models also show relatively high rates, while several open-source models in LMSYS show lower rates (such as \texttt{alpaca-13b}, \texttt{llama-2-13b-chat}, \texttt{vicuna-13b}, etc.). However, these patterns should not be interpreted as causal model rankings, since models differ in dataset coverage, user populations, prompt topics, deployment and time periods. We therefore report the full model-name table in Appendix~\ref{app:model_curiosity} as descriptive evidence that curiosity-oriented questioning can vary across model identities.

\subsection{Model Analysis}
\label{app:model_curiosity}

\begin{table}[t]
\centering
\tiny
\begin{tabular}{llr}
\toprule
Model & Dataset(s) & Curiosity rate \\
\midrule
gpt-4 & WildChat, LMSYS & 54.18\% \\
claude-2 & LMSYS & 47.90\% \\
claude-1 & LMSYS & 46.22\% \\
claude-instant-1 & LMSYS & 42.52\% \\
gpt-3.5-turbo & WildChat, LMSYS & 41.43\% \\
vicuna-33b & LMSYS & 34.22\% \\
guanaco-33b & LMSYS & 27.39\% \\
oasst-pythia-12b & LMSYS & 26.89\% \\
mpt-30b-chat & LMSYS & 26.66\% \\
llama-13b & LMSYS & 26.26\% \\
palm-2 & LMSYS & 25.96\% \\
mpt-7b-chat & LMSYS & 25.31\% \\
alpaca-13b & LMSYS & 24.57\% \\
stablelm-tuned-alpha-7b & LMSYS & 24.33\% \\
wizardlm-13b & LMSYS & 23.01\% \\
llama-2-13b-chat & LMSYS & 22.89\% \\
RWKV-4-Raven-14B & LMSYS & 22.92\% \\
vicuna-13b & LMSYS & 21.84\% \\
koala-13b & LMSYS & 21.58\% \\
chatglm-6b & LMSYS & 20.85\% \\
llama-2-7b-chat & LMSYS & 19.76\% \\
\bottomrule
\end{tabular}
\caption{Curiosity rate by explicit model name. Rates are computed over real, classified assistant questions only. Dataset-level
placeholders such as \texttt{oasst1} and \texttt{sharegpt} are excluded.}
\label{tab:model_curiosity}
\end{table}

Table~\ref{tab:model_curiosity} shows model-wise variation of curiosity rates across datasets. 

For each model, we compute a normalized curiosity rate over real, classified LLM questions, defined as \textbf{\(\#\text{curiosity}/(\#\text{curiosity}+\#\text{non-curiosity})\)}. We exclude artifacts, unclear cases, and problematic cases from the above computation. We observe substantial variation across models: \texttt{gpt-4} has a higher curiosity rate than \texttt{gpt-3.5-turbo}, Claude models also show relatively high rates, while several open-source models in LMSYS show lower rates (such as \texttt{alpaca-13b}, \texttt{llama-2-13b-chat}, \texttt{vicuna-13b}, etc.). However, these patterns should not be interpreted as causal model rankings, since models differ in dataset coverage, user populations, prompt topics, deployment and time periods. We therefore report the full model-name table in Appendix~\ref{app:model_curiosity} as descriptive evidence that curiosity-oriented questioning can vary across model identities.

\section{Q2-Methods}

\subsection{User-Side Curiosity and Engagement Proxies}
\label{app:q2_human_proxy}

 RQ2 asks whether model curiosity is associated with richer downstream user behavior. Because the datasets are large-scale public conversation logs, we cannot directly observe users' internal curiosity, motivation, or engagement. We therefore operationalize user-side outcomes as observable conversational behaviors in the next user turn. These measures should be interpreted as scalable proxies rather than direct psychological measurements.

\paragraph{Rule-based engagement proxies.} We compute several transparent textual indicators from the user turn following the model response. \textit{Any question} records whether the next user turn contains at least one question. \textit{Exploratory questions} identify question forms that invite explanation, comparison, alternatives, implications, causes, or possibilities, rather than only short factual lookup. \textit{Epistemic markers} count lexical cues associated with reasoning or knowledge-state management, such as uncertainty, evidence, explanation, comparison, causality, and reflection. \textit{Response length} measures the number of words in the next user turn. Together, these proxies capture whether the user continues the interaction in a way that is more elaborated, question-asking, or reasoning-oriented.

\paragraph{LLM-judged user curiosity.} In addition to surface proxies, we use an LLM judge to label whether the subsequent user turn expresses curiosity. The judge is given the relevant conversational context and labels only curiosity expressed in the user's own following turn, not curiosity inferred from the model's preceding question. A user turn is labeled curiosity-expressing when it explicitly seeks or develops understanding, explanation, evidence, alternatives, implications, perspectives, or reflection. This can appear as a direct question, an indirect request for explanation, or an imperative such as asking the model to explain, compare, justify, or explore an issue further. This uses the taxonomy developed in RQ1. 

\noindent \textbf{Human validation of downstream curiosity.} We conduct an additional human validation of LLM-judged downstream user curiosity using two annotators. Human judgments are compared against the LLM classifications, leading to a 89.9\% agreement
and Cohen's $\kappa = 0.838$.

\paragraph{Interpretation.} These outcomes are intended to measure observable downstream behavior, not latent mental states.  We therefore report the RQ2 results as adjusted associations rather than causal estimates of users' internal curiosity.

We use two complementary analyses for the above: \textbf{(1) matched analysis} and\textbf{ (2) adjusted regression}. In both, the treatment is a curiosity-oriented model question and the control is a non-curiosity model question. We exclude contexts where the user explicitly asked the model to generate questions.

\subsection{Modeling}
\label{app:main_q2}

For matched analysis, we pair each curiosity-oriented model question with a non-curiosity model question from the same model family and broad topic category. These matched pairs belong to the same model family and broad topic category. Within these blocks, we select the nearest control using pre-treatment features: prior user length, question count, epistemic markers, exploratory questions, assistant response length, turn index, and conversation length. We then compare the next user turn for these matched pairs. We use this design because large-scale naturally occurring conversations are difficult to randomize directly. Stronger causal evidence would require controlled human studies that manipulate model curiosity. We therefore interpret matched differences as adjusted associations and residual confounding may remain from unobserved context (See Appendix~\ref{app:matched} for more details). 

Within these exact blocks, we select the nearest control using pre-treatment conversational features: previous user word count, previous user question count, previous user epistemic marker count, previous user exploratory question count, assistant response length, assistant turn index, and total conversation length. We then compare the next user turn across matched pairs. This design reduces confounding. However, this design does not claim that curiosity and non-curiosity turns are identical. Rather, it constructs a more comparable observational contrast by matching on information available before the downstream user response. Exact matching on model family and broad topic keeps pairs within similar conversational settings, while nearest-neighbor matching reduces imbalance in prior user behavior, assistant verbosity, and conversation position. We therefore interpret matched differences as adjusted associations, not causal effects. Residual confounding may remain if curiosity questions occur in contexts that are not fully captured by these observed features.

For the regression analysis, we fit adjusted models over all eligible assistant-question events. For each downstream outcome, we
regress the next-user measure on a binary curiosity indicator while controlling for the same pre-treatment features used in matching,
as well as model-family and topic-category fixed effects. The coefficient on the curiosity indicator estimates the adjusted association between curiosity-oriented questioning and the next user response. Both analyses are observational and should be interpreted as correlational rather than causal.

To validate the rule-based downstream measures, we annotate a matched sample of next-user turns with an independent LLM judge. For each matched pair, the judge sees the previous user turn, the assistant question, and the subsequent user turn, and labels whether the subsequent user turn expresses human curiosity, its curiosity type, and whether non-curiosity turns are mainly task continuation, acknowledgment, formatting, or answering the assistant. This validation is applied to 2000 matched pairs per dataset where available, with all 468 OASST1 pairs included.

\subsection{Matched Analysis}
\label{app:matched}

For the matched analysis, we match each curiosity question to a non-curiosity question from the same model family and broad topic category. Within these exact blocks, we select the nearest control using pre-treatment conversational features: previous user word count, previous user question count, previous user epistemic marker count, previous user exploratory question count, assistant response length, assistant turn index, and total conversation length. We then compare the next user turn across matched pairs. This design reduces confounding. However, this design does not claim that curiosity and non-curiosity turns are identical. Rather, it constructs a more comparable observational contrast by matching on information available before the downstream user response. Exact matching on model family and broad topic keeps pairs within similar conversational settings, while nearest-neighbor matching reduces imbalance in prior user behavior, assistant verbosity, and conversation position. We therefore interpret matched differences as adjusted associations, not causal effects. Residual confounding may remain if curiosity questions occur in contexts that are not fully captured by these observed features.

 \paragraph{Why matched observational analysis?} Our goal is to estimate whether curiosity-oriented model questions are associated with richer downstream user behavior in large-scale, naturally occurring conversations. A fully causal design would require randomly assigning users or prompts to receive curiosity-oriented versus non-curiosity model responses and then measuring subsequent engagement. Such a design is difficult to implement at the scale and diversity of open-ended conversation logs, and it would introduce its own confounds, including participant expectations, prompt selection effects, task familiarity, selected using pre-treatment features such as prior user length, prior question count, epistemic markers, exploratory questions, assistant response length, turn index, and conversation length. This design reduces observable imbalance before the downstream user response, but it cannot eliminate unobserved confounding or establish causality. We interpret the resulting estimates as adjusted associations and use them to motivate, rather than replace, future controlled human experiments.

\subsection{Regression}

\subsubsection{Regression Significance Testing}
\label{app:q2-significance}

For regression-based estimates, we compute two-sided significance tests using robust standard errors clustered at the conversation level, so that multiple turns from the same conversation are not treated as independent evidence. For matched validation estimates, including the LLM-judge human-curiosity estimate, we use paired bootstrap resampling over matched treatment-control pairs. Each bootstrap replicate samples matched pairs with replacement and recomputes the treatment-control difference, yielding percentile confidence intervals. We report significance markers in Table~\ref{tab:q2-regression-llm-results} as descriptive uncertainty summaries. Because treatment assignment is observational, statistical significance should not be interpreted as causal identification; it indicates that the adjusted association is unlikely to be explained by sampling variability alone under the fitted adjustment design.

\subsection{LLM Judge Curiosity}
As a validation analysis, we use an independent LLM judge to classify downstream human curiosity in the matched Q2 sample. For each treatment-control pair, we construct two annotation items: the next-user turn following the curiosity-oriented assistant question and the next-user turn following the matched non-curiosity control question. The judge receives limited local context consisting of the previous user turn, the assistant question, the assistant-question type, and the next user turn to be labeled. It is instructed to label only the next user turn, not the assistant question. The output schema includes a top-level status (\texttt{human\_curiosity}, \texttt{not\_human\_curiosity}, or \texttt{unclear}), a human-curiosity subtype (\texttt{information\_seeking}, \texttt{causal\_explanatory}, \texttt{exploratory\_broadening}, \texttt{reflective\_metacognitive}, \texttt{perspective\_taking}, \texttt{counterfactual\_hypothetical}, or \texttt{none}), a non-curiosity function for negative cases, a 0-3 curiosity-intensity score, confidence, and a short explanation. We run this validation on 2000 matched pairs from WildChat, LMSYS, and ShareGPT, and on all 468 matched pairs from OASST1. This design keeps the primary Q2 outcomes transparent and reproducible, while testing whether those surface proxies align with a semantic judgment of downstream human curiosity.

\section{Q2-Results}
\begin{figure}
\centering
\includegraphics[width=\linewidth]{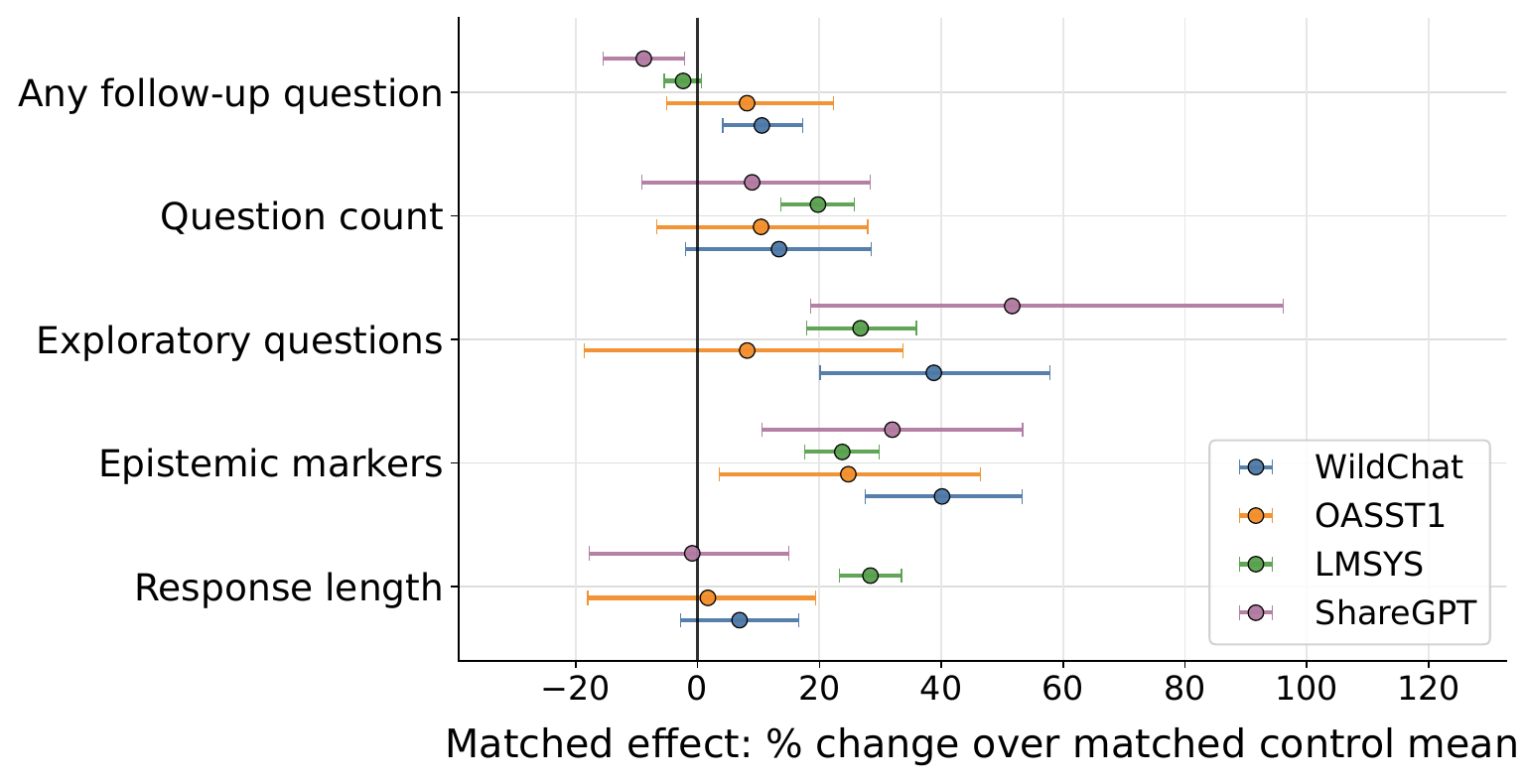}
\caption{Matched Effects Analysis - Stage 1. }
\vspace{-10.5pt}
\label{fig:matched_stage1}
\end{figure}


\begin{figure}
\centering
\includegraphics[width=\linewidth]{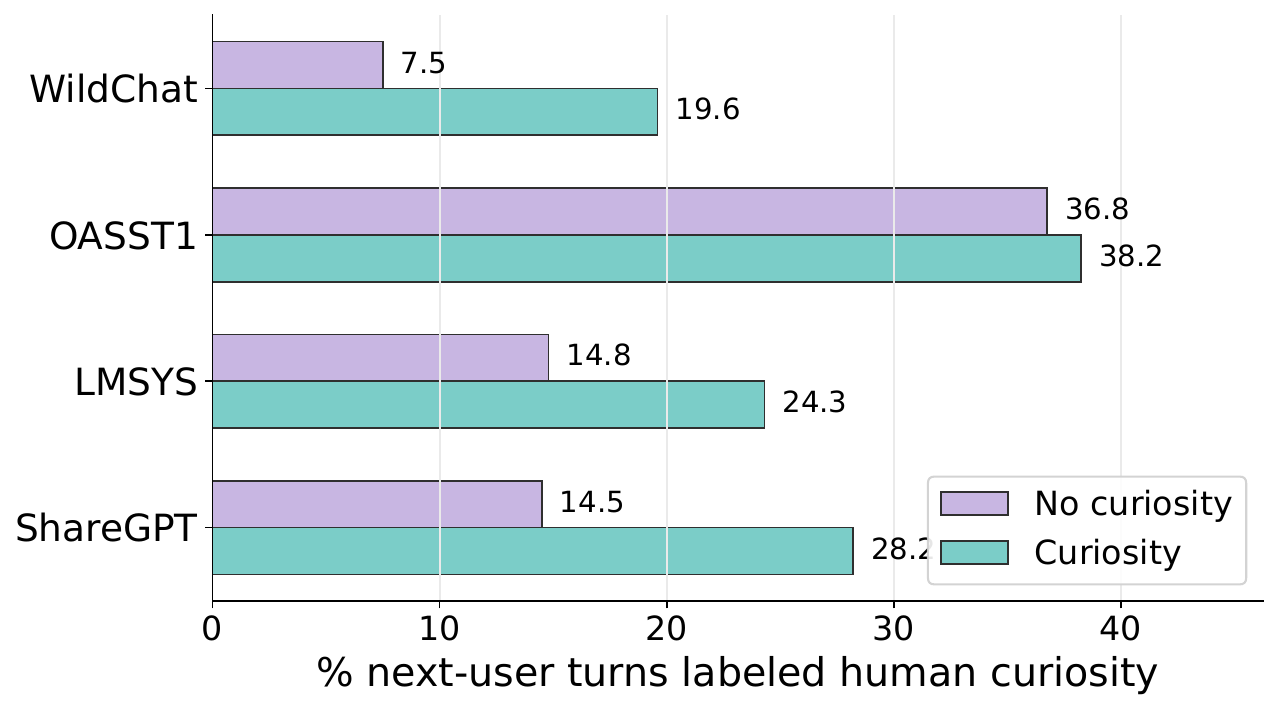}
\caption{Curiosity Rates - Stage 2. }
\vspace{-10.5pt}
\label{fig:matched_stage2}
\end{figure}

\begin{figure}
\centering
\includegraphics[width=\linewidth]{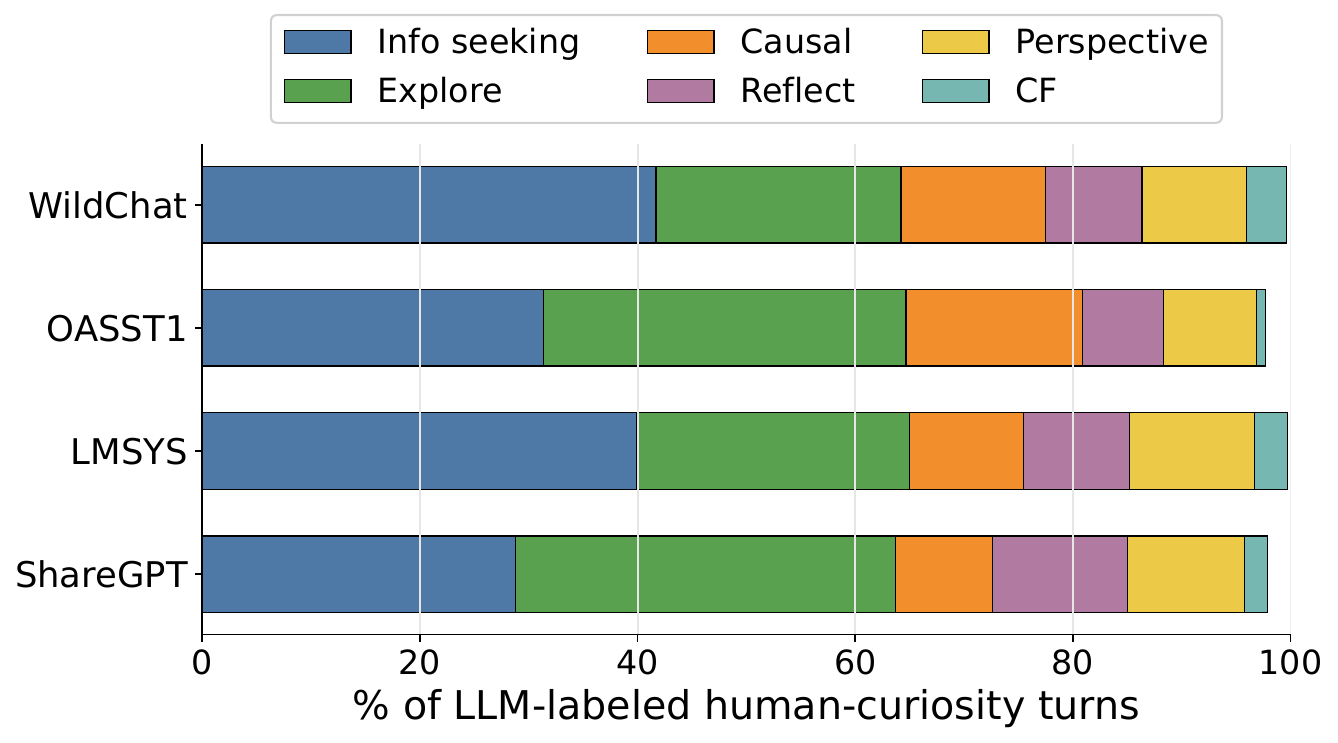}
\caption{Curiosity Types by User - Stage 2. }
\vspace{-10.5pt}
\label{fig:matched_stage2_}
\end{figure}

\subsection{Causal and Longitudinal Analyses- methodology and additional experiments}
\label{app:q2-causal-method}
We complement the matched and regression analyses with an
inverse-probability-weighted estimate of the average treatment effect on
the treated (ATT). Let $T_i=1$ denote a curiosity-oriented model question,
$T_i=0$ a clean non-curiosity question, and $Y_i$ the subsequent user-turn
outcome. We estimate the propensity score
$\hat{e}(X_i)=P(T_i=1\mid X_i)$ using previous-user word count, question
count, exploratory-question count, epistemic-marker count, assistant-turn
length and position, and conversation length. Treated observations receive
weight 1, while controls receive weight
$\hat{e}(X_i)/(1-\hat{e}(X_i))$. The normalized estimator is

\[
\widehat{\tau}_{\mathrm{ATT}}
=
\frac{1}{N_1}\sum_{i:T_i=1}Y_i
-
\frac{
\sum_{i:T_i=0}
\frac{\hat{e}(X_i)}{1-\hat{e}(X_i)}Y_i
}{
\sum_{i:T_i=0}
\frac{\hat{e}(X_i)}{1-\hat{e}(X_i)}
}.
\]

The immediate matched validation uses the strict human-curiosity outcome, whereas the longitudinal analysis uses the broader human-curiosity-presence outcome, judged by \texttt{gpt-4o-mini}, across $t{+}1$--$t{+}5$. IPW is applied only to immediate rule-based engagement outcomes from the original event-level file; the separately sampled multi-turn trajectories are used only for the longitudinal LLM-judge analysis. The rule-based engagement and LLM-judged human-curiosity outcomes are therefore reported and interpreted independently.

We report ATT estimates as a sensitivity analysis rather than as identified
causal effects. For the subtype analysis, each specific curiosity subtype
is compared with the common clean-control group, with 95\%
conversation-clustered confidence intervals. For the trajectory analysis,
we sample complete treatment and control trajectories and independently
judge human curiosity (not just engagement) in user turns from $t{+}1$ to $t{+}5$.
Roleplay, generated artifacts, and turns that are not genuine user
responses are excluded. 

For the additional experiments: specifically Extension 3 (refer to ~\ref{app: extension3-causalanalysis})  and Extension 4 (refer to ~\ref{app: extension4-causalanalysis}), within the 25,948 events assigned to nine named prompt-context categories, deeper curiosity subtypes remain uncommon; the remaining 78.7\% of clean events are labelled \textit{other} and are not shown in this stratification. Prior-user-curiosity ATT point estimates are largest among users in the high-curiosity tercile across all four datasets. This descriptive pattern is consistent with model curiosity amplifying existing epistemic engagement, but does not show that such engagement is causally initiated by the model.

Together, the immediate-turn rule-based measures and LLM judgments associate model curiosity with richer epistemic content and higher judged human-curiosity rates, although the magnitude varies across datasets, outcomes, and adjustment methods.

\subsection{Covariate Balance for Matched RQ2 Analysis}
\label{app:q2_covariate_balance}

To assess comparability of matched treatment and control turns, we report standardized mean differences (SMDs) for the pre-treatment covariates used in matching: prior user length, prior question count, prior epistemic markers, prior exploratory questions, assistant response length, assistant turn index, and conversation length. Matching constrains comparisons to the same model family and broad topic category, but some residual imbalance remains, especially for assistant response length in WildChat, OASST1, and ShareGPT, and prior user length in OASST1 and LMSYS. We therefore include the same covariates in adjusted regression analyses and interpret all estimates as adjusted associations rather than causal effects.

  \begin{table*}[t]
  \centering
  \small
  \setlength{\tabcolsep}{3.5pt}
  \begin{tabular}{lrrrrr}
  \toprule
  \textbf{Covariate} & \textbf{WildChat} & \textbf{OASST1} & \textbf{LMSYS} & \textbf{ShareGPT} & \textbf{All} \\
  \midrule
  Prior user words & 0.278 & 0.516 & 0.507 & 0.187 & 0.338 \\
  Prior questions & 0.140 & 0.107 & 0.197 & 0.166 & 0.138 \\
  Prior epistemic markers & 0.274 & 0.481 & 0.316 & 0.193 & 0.254 \\
  Prior exploratory questions & 0.237 & 0.234 & 0.186 & 0.154 & 0.180 \\
  Assistant words & 0.809 & 0.718 & -0.023 & 0.640 & -0.008 \\
  Assistant turn index & 0.154 & -0.002 & 0.024 & -0.080 & 0.036 \\
  Conversation turns & 0.053 & -0.038 & 0.006 & -0.072 & 0.036 \\
  \bottomrule
  \end{tabular}
  \caption{Standardized mean differences between matched curiosity and non-curiosity events on pre-treatment covariates. Positive values indicate
  higher covariate values for curiosity-oriented treatment turns; negative values indicate higher values for matched controls.}
  \label{tab:q2_covariate_balance}
  \end{table*}

\subsection{Additional Causal Analysis Experiments and Results}
\label{app:q2-causal-extensions}

\begin{figure}[ht]
\centering
\includegraphics[width=\linewidth]{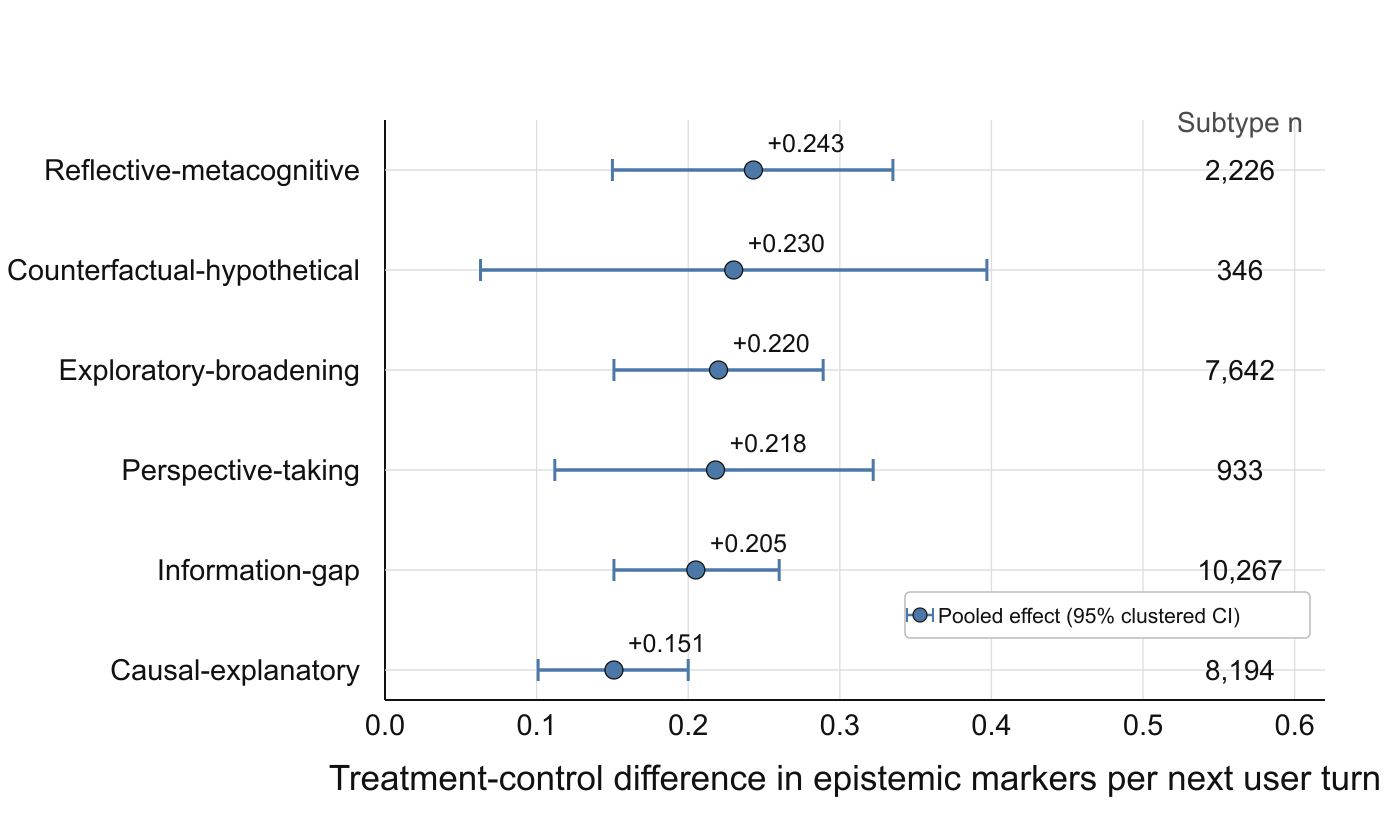}
\caption{Subtype-specific differences in mean next-user epistemic-marker count relative to clean controls. Error bars show 95\% conversation-clustered intervals.}
\label{fig:q2-subtype-effects}
\end{figure}

\subsubsection{Curiosity Subtype.}
\label{app:causal_subtype}

We study whether the next user turn exhibits different levels of epistemic behavior depending on the subtype of curiosity expressed in the preceding model question. \textbf{Reflective-metacognitive questions show the largest pooled increase in next-turn epistemic markers (+0.243), followed by counterfactual-hypothetical (+0.230) and exploratory-broadening questions (+0.220). Causal-explanatory questions show the smallest increase (+0.151), as seen in Figure ~\ref{fig:q2-subtype-effects}}. The ordering varies across datasets, while the counterfactual-hypothetical and perspective-taking estimates are based on smaller treatment samples.

\subsubsection{Multi-turn Persistence Longitudinal Analysis}
\label{app:q2-causal-extensions-q2}
\begin{figure}[t]
\centering
\includegraphics[
    width=\linewidth,
    height=0.23\textheight,
    keepaspectratio
]{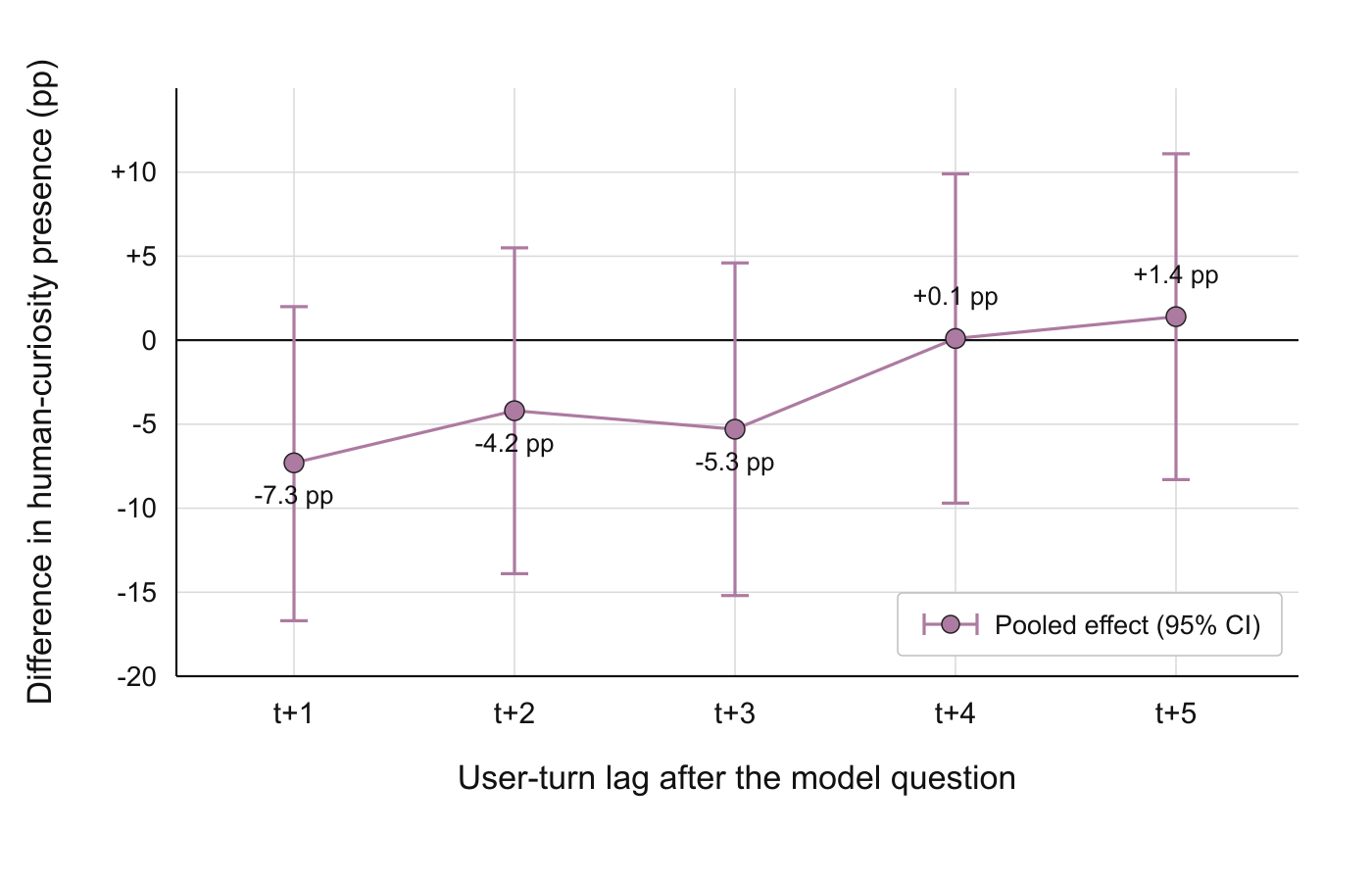}
\caption{Pooled treatment-control differences in judged human-curiosity presence across five later user turns}
\label{fig:q2-multiturn-engagement}
\end{figure}

\begin{figure}[t]
\centering
\includegraphics[width=\linewidth]{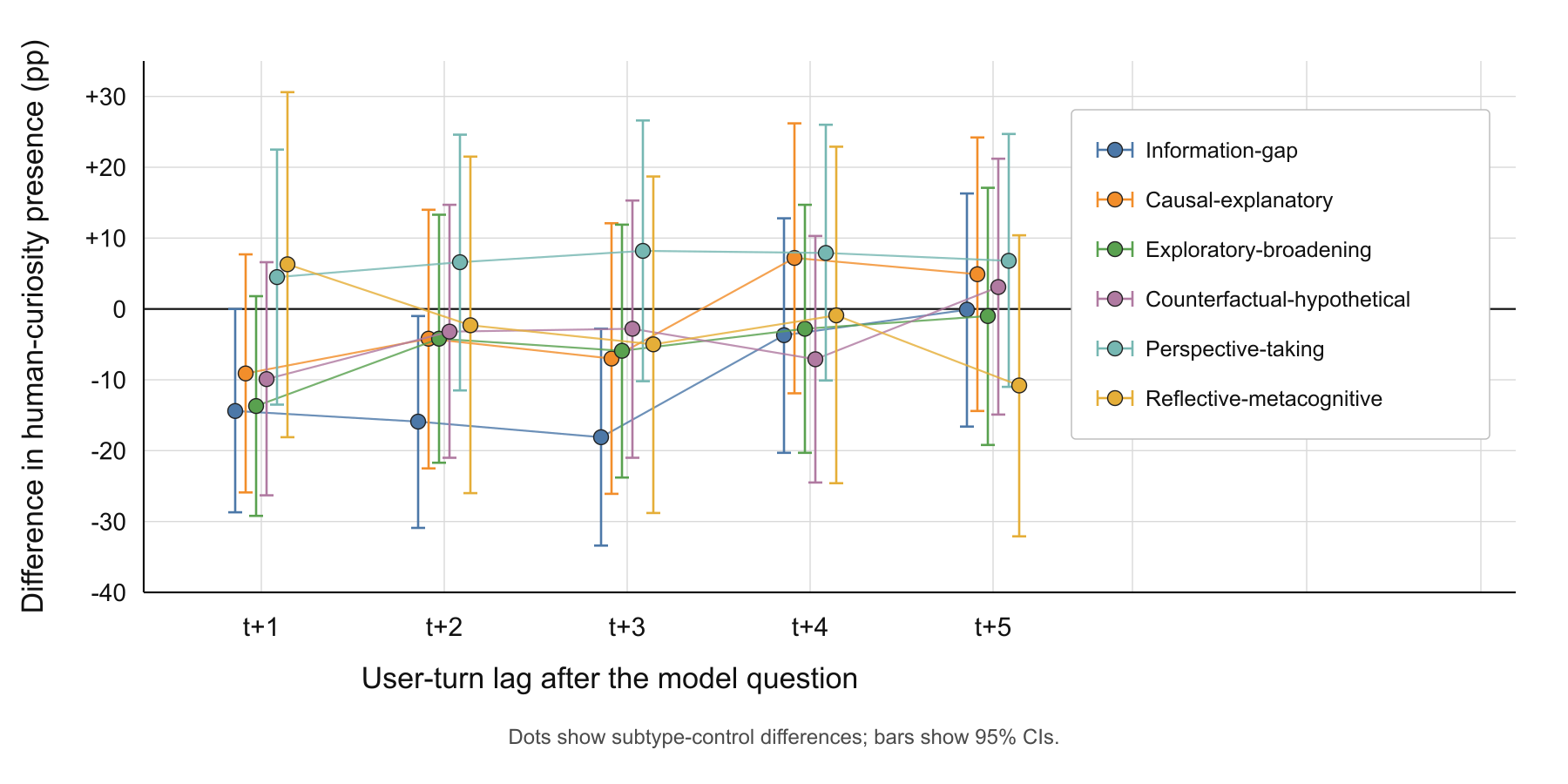}
\caption{Subtype-control differences in judged human-curiosity presence across five later user turns. Points show rate differences and bars show 95\% confidence intervals.}
\label{fig:q2-subtype-trajectories}
\end{figure}

\paragraph{Multi-turn persistence and delayed human curiosity emergence analysis}
Although the trajectory is non-monotonic and all 95\% confidence intervals include zero, it indicates a change in timing rather than an absence of association. The curiosity-question group trails the control group from $t{+}1$ to $t{+}3$, reaches approximately the same level at $t{+}4$ ($+0.1$ pp), and exceeds it at $t{+}5$ ($+1.4$ pp) as seen in ~\ref{fig:q2-multiturn-engagement}.This crossover is consistent with relative persistence because human curiosity remains relatively stable after curiosity-oriented questions while declining more quickly in controls, suggesting that model-elicited curiosity may persist longer and become visible only after several exchanges. The subtype results reinforce this interpretation: perspective-taking, causal-explanatory, and counterfactual-hypothetical questions are positive at $t{+}5$ as seen in ~\ref{fig:q2-subtype-trajectories}, whereas direct information-gap questions converge near zero as their gaps are resolved. We therefore interpret the $t{+}5$ positivity as descriptive evidence of a delayed, subtype-dependent human-curiosity response, while treating its precise magnitude cautiously.

Here are two valid treatment trajectories illustrating the re-emergence of judged human curiosity at $t{+}5$. Conversation text is paraphrased for brevity.

\textbf{Example 1: Meritocracy and inequality (LMSYS)- Perspective taking.}
The model initially asked whether education-based meritocracy increases or reduces inequality. After several intervening turns, the user returned to the concept at $t{+}5$ by asking whether schools themselves subscribe to meritocratic ideology. Judged human curiosity was present through $t{+}3$, absent at $t{+}4$, and present again at $t{+}5$.

\textbf{Example 2: Subjective and objective reality (ShareGPT)- Perspective taking.}
The model introduced the distinction between subjective and objective reality and invited the user's position. At $t{+}5$, the user revisited the issue by requesting a structured counterargument against objective reality. Judged human curiosity was again present through $t{+}3$, absent at $t{+}4$, and present at $t{+}5$, illustrating delayed re-engagement with the model's earlier question.

\paragraph{Subtype-specific trajectories} 
The judge identifies human curiosity through observable epistemic behavior, including information seeking, explanation seeking, exploration, and reflection; it does not assign human responses the model-side curiosity taxonomy. We disaggregate the trajectory analysis by the subtype of the source model question. Figure~\ref{fig:q2-subtype-trajectories} reports each subtype's difference from pooled clean controls in judged human-curiosity presence across five lags. Perspective-taking questions show consistently positive point estimates from $t{+}1$ to $t{+}5$ ($+4.5$ to $+8.2$ percentage points). Information-gap questions are lower than controls through $t{+}3$ but converge to near zero by $t{+}5$, while causal-explanatory questions become descriptively positive at $t{+}4$ and $t{+}5$. Most confidence intervals cross zero; reflective-metacognitive estimates are particularly uncertain because only 17 valid treatment turns are available per lag. These subtype patterns are therefore descriptive, especially given the multiple subtype-lag comparisons.

\subsubsection{Curiosity-Triggering Context}
\label{app: extension3-causalanalysis}
\begin{figure}[t]
\centering
\includegraphics[width=\linewidth]{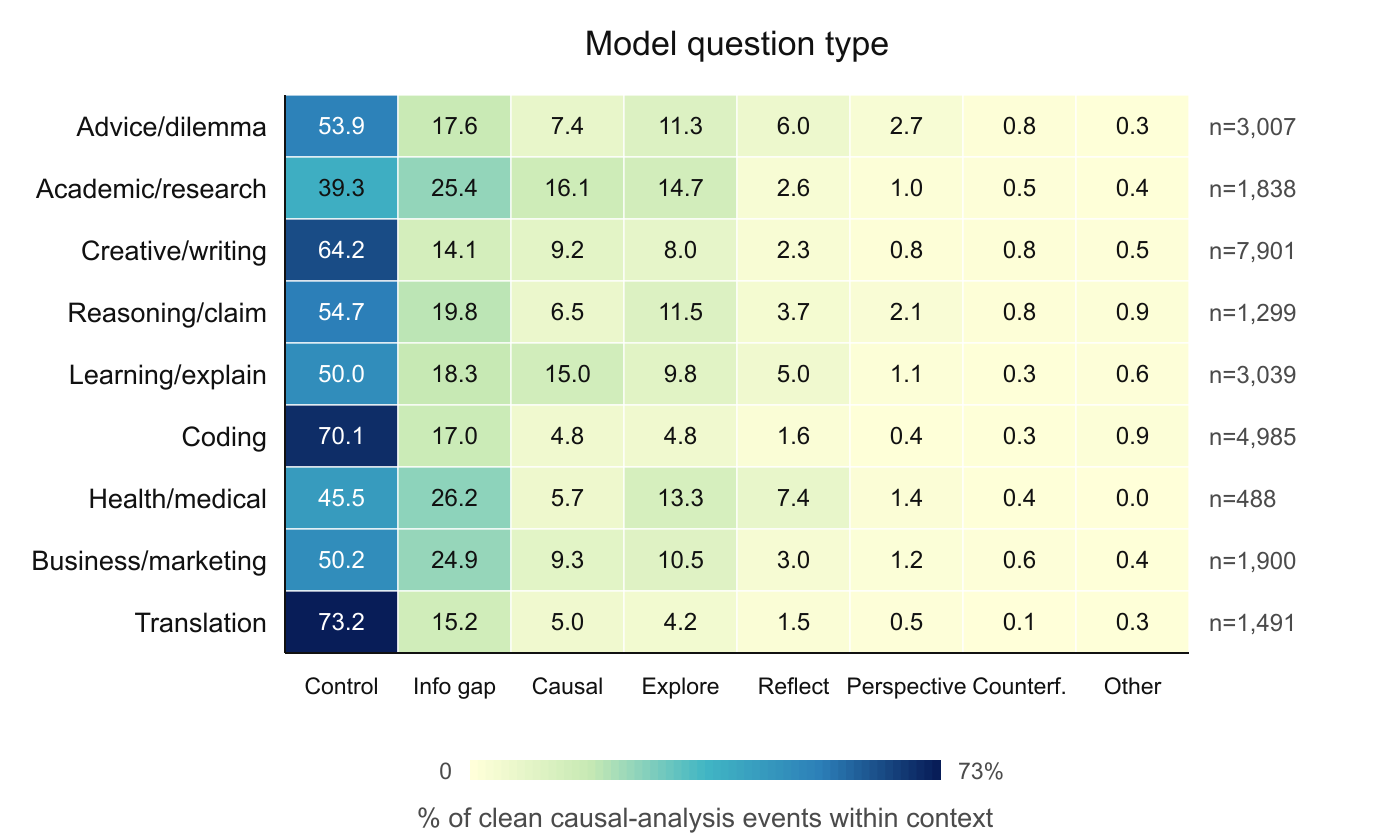}
\caption{Model-question composition within the nine named preceding-user contexts in the clean causal-analysis event table.}
\label{fig:q2-context-composition}
\end{figure}

We stratify the clean causal-analysis events by the preceding user context. The nine displayed context categories contain 25,948 events, corresponding to 21.3\% of the 121,669-event clean sample; the remaining 95,721 events are labeled \textit{other} and are not shown. Within the named contexts, clean non-curiosity controls are the most common category, while reflective-metacognitive, counterfactual-hypothetical, and perspective-taking questions remain rare. Question-type composition varies descriptively across contexts, but models rely primarily on information-gap, causal-explanatory, and exploratory-broadening questions rather than deeper forms of inquiry. Refer to Figure ~\ref{fig:q2-context-composition}

\subsubsection{Prior User Engagement and Epistemic Downstream Behavior}
\label{app: extension4-causalanalysis}
\begin{figure}[t]
\centering
\includegraphics[width=\linewidth]{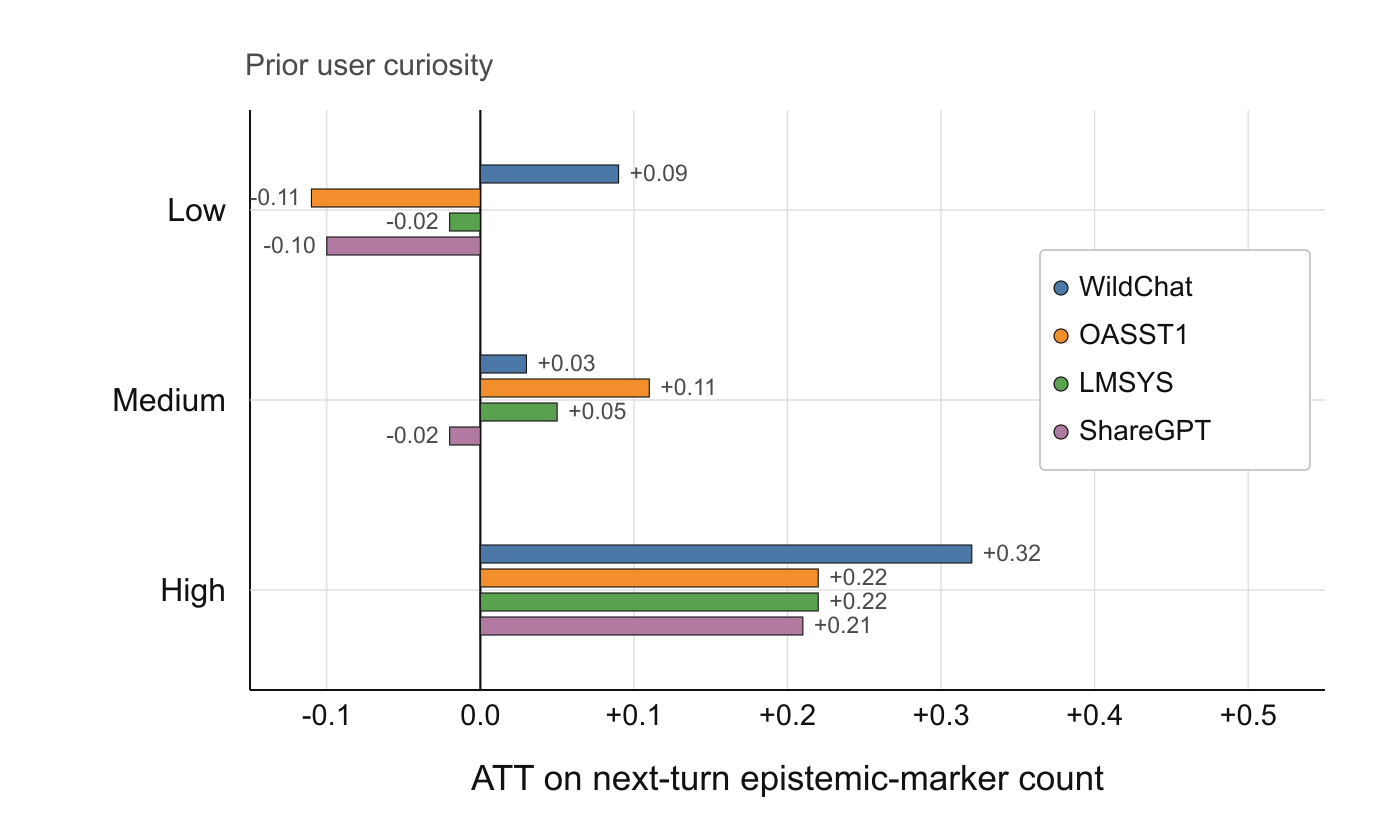}
\caption{ATT point estimates for next-turn epistemic markers by prior-user-engagement tercile and dataset. }
\label{fig:q2-prior-curiosity}
\end{figure}

We divide users into dataset-specific terciles based on epistemic behavior in the preceding turn and report ATT point estimates for epistemic markers in the next-turn. Across all four datasets, the largest estimate occurs among users in the high-engagement tercile, ranging from +0.21 to +0.32, as seen in Figure ~\ref{fig:q2-prior-curiosity}. Estimates for low- and medium-engagement users are smaller and mixed. This pattern is consistent with model curiosity amplifying an existing epistemic orientation rather than reliably initiating one. Because uncertainty intervals are unavailable for this stratified analysis, the estimates are interpreted descriptively rather than as evidence of statistically significant moderation. Importantly, these terciles represent the user's immediate conversational state rather than a stable personal trait. The results suggest that curiosity-oriented questions show the largest estimates when they build on an already active exploratory goal. Consequently, LLMs may benefit in varying whether and how they ask questions according to the user's current epistemic engagement.

\subsubsection{Causal analysis results as empirical evidence for RQ3.} Our RQ2 causal analyses naturally provide the empirical basis for RQ3. Extension 1 identifies which curiosity strategies are associated with larger immediate differences, Extension 2 reveals which strategies sustain those benefits over multiple conversational turns, and Extension 3 characterizes the contexts in which different strategies are currently deployed. Together, these analyses provide evidence-based criteria for defining useful curiosity and identifying missed opportunities.

\section{Q3 - Methods and Results}

\subsection{Human Study}
\label{app:human_study_q3}

Participants completed the study on Prolific. After consent and brief instructions, each participant was shown a randomized subset of items. Each item contained the original user message and three candidate assistant responses: a direct answer, a non-curiosity follow-up question, and a curiosity-oriented follow-up question. Participants then answered preference-style annotation questions about the responses, including which response they preferred, which felt most natural, which would make them more curious or likely to continue, and whether the curiosity-oriented question was helpful, optional, or unnecessary. We also asked whether the curiosity-oriented question felt annoying, burdensome, invasive, or unsafe. Finally, we asked them if a curiosity question should/could be in the response at all. The order and block assignment were randomized so that each participant annotated only a portion of the full item set. We included an attention check and excluded responses from participants who failed the attention check, completed no substantive items, or selected the same option for every preference item. Fig~\ref{fig:prolific-guidelines} illustrates these guidelines. 

\begin{figure}[ht]
\centering
\begin{tcolorbox}[
    colback=blue!5,
    colframe=blue!70!black,
    title=\textbf{Prolific Human Study Annotation Guidelines},
    fonttitle=\bfseries,
    arc=2mm,
    boxrule=0.8pt
]
\scriptsize
Participants read each user message and compared three possible assistant responses: \textbf{(A)} a direct answer, \textbf{(B)} a non-curiosity follow-up question, and \textbf{(C)} a curiosity-oriented follow-up question. They judged the responses from the user's perspective, focusing on what would be most useful, natural, or engaging in context.

For each item, participants answered:

\begin{enumerate}[leftmargin=*]
    \item \textbf{Preferred response:} Which assistant response would you prefer?
    \item \textbf{Most natural response:} Which assistant response feels most natural?
    \item \textbf{Curiosity or continuation:} Which response would make you more curious or more likely to continue?
    \item \textbf{Usefulness of the curiosity question:} Is the curiosity-oriented question helpful, optional, or unnecessary?
    \item \textbf{Potential burden:} Does the curiosity-oriented question feel annoying, burdensome, invasive, or unsafe?
\end{enumerate}

Participants were told that there were no universally correct answers and to judge each item based on the specific user message and candidate responses.
\end{tcolorbox}
\caption{Annotation guidelines for the Prolific human study on useful model curiosity.}
\label{fig:prolific-guidelines}
\end{figure}

\begin{figure}[ht]
\centering
\begin{tcolorbox}[
    colback=green!5,
    colframe=green!60!black,
    title=\textbf{Example Prolific Annotation Item},
    fonttitle=\bfseries,
    arc=2mm,
    boxrule=0.8pt
]
\scriptsize
\textbf{User message:} I am thinking about changing careers, but I am worried I might regret leaving my current job.

\vspace{3pt}
\textbf{Choice A: Direct answer}\\
Changing careers can be a good decision if your current role no longer fits your goals. You should compare your financial stability, long-term interests, skills, and available opportunities before deciding.

\vspace{3pt}
\textbf{Choice B: Non-curiosity question}\\
That is a tough one. Would you like me to give a straightforward answer?

\vspace{3pt}
\textbf{Choice C: Curiosity-oriented question}\\
What parts of your current job feel hardest to give up, and what are you hoping a new career would make possible?

\vspace{4pt}
\textbf{Annotation questions:}
\begin{enumerate}[leftmargin=*]
    \item Which model response would you prefer? \texttt{Choice A / Choice B / Choice C}
    \item Which response feels most natural? \texttt{Choice A / Choice B / Choice C}
    \item Which response would make you more curious or more likely to continue? \texttt{Choice A / Choice B / Choice C}
    \item Is the curiosity-oriented question helpful here? \texttt{Yes / Optional / No}
    \item Does the curiosity-oriented question feel annoying, burdensome, invasive, or unsafe? \texttt{Yes / No}
\end{enumerate}
\end{tcolorbox}
\caption{Example item from the Prolific human study comparing a direct answer, a non-curiosity question, and a curiosity-oriented question.}
\label{fig:prolific-example-item}
\end{figure}

Finally, participants were compensated according to Prolific's fair-payment guidelines.

\noindent \textbf{Candidate Questions for Survey.} We construct survey items from candidate missed opportunities: user queries that contain uncertainty, learning goals, advice dilemmas, creative or research framing, reasoning gaps, or other antecedents identified in Fig.~\ref{fig:antecedent}, but where the observed model response does not ask a follow-up question. We retain cases that are not followed by an LLM curiosity question, since these mark plausible moments where curiosity may have been useful but absent. To test the boundary between useful and unnecessary curiosity, we also include contrast cases where a direct answer may be sufficient.

For each selected user query, we use \texttt{gpt-5} to generate two candidate follow-up questions: one curiosity-oriented question, guided by our curiosity taxonomy, and one non-curiosity question, such as a task-clarification, engagement-maintenance, or workflow-management question. These generated questions are used only to construct controlled survey alternatives; their usefulness is not assumed. The human preference study then evaluates whether participants prefer the direct answer, the non-curiosity question, or the curiosity-oriented question, and whether the curiosity-oriented question is judged helpful, optional, burdensome, or unnecessary in context. Fig~\ref{fig:prolific-example-item} shows an example of an item used for the human study. 

\noindent \textbf{Presentation controls.} The response alternatives were semantically fixed to preserve the comparison: Choice A was always the direct answer, Choice B the non-curiosity question, and Choice C the curiosity-oriented question. This makes the contrast interpretable, but it also means that labels were not position-randomized. To reduce presentation confounds, the two question alternatives were generated under the same item-level prompt and constrained to be short follow-up questions. We audit response lengths and manually inspect generated questions for obvious quality failures. Because participants may still be sensitive to style, specificity, or length differences across options, we interpret preference and naturalness judgments as contextual usefulness signals rather than pure estimates of curiosity alone.

\subsubsection{Human Study Agreement}
\label{app:human_study_agreement}

We summarize annotation consistency using item-level modal agreement, since each item is rated by only a small number of participants. Agreement is 73.5\%--77.4\% for questions where participants select among the direct answer, the non-curiosity question, and the curiosity-oriented question. Agreement is lower for whether the model should ask the curiosity question (\textit{Yes}/\textit{Optional}/\textit{No}; 61.9\%), reflecting that participants often differ on whether a follow-up question is necessary or merely optional. Agreement is also overall modest because it asks for subjective preference judgments and also sample size is smaller.

For the useful-curiosity analysis, we also derive binary labels from the survey responses. These labels indicate whether the curiosity-oriented question was selected as the preferred response, selected as the most curiosity-awakening response, selected as the most natural response, judged acceptable to ask (\textit{Yes} or \textit{Optional}), or judged burdensome/annoying. The final useful-curiosity label combines these signals to identify cases where the curiosity-oriented question received positive support without strong burden. Modal agreement for these derived binary labels ranges from 75.0\% to 84.2\%.

\subsubsection{Demographic Differences}
\begin{figure}[t]
\centering
\includegraphics[width=\linewidth]{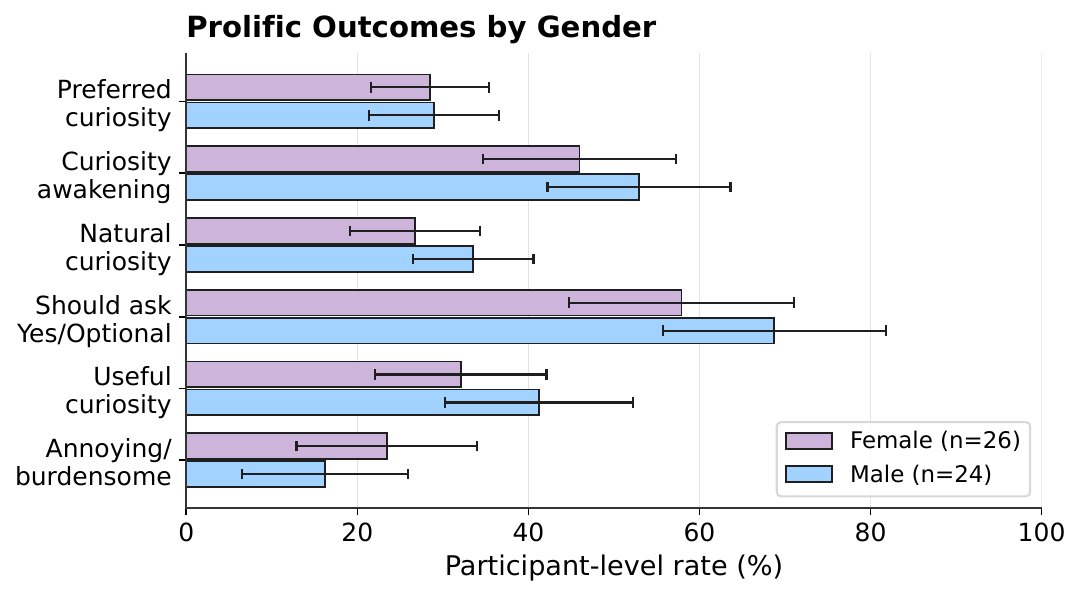}
\caption{Gender-Based Differences in Prolific Study for Q3}
\label{fig:q3-gender}
\end{figure}

\begin{figure}[t]
\centering
\includegraphics[width=\linewidth]{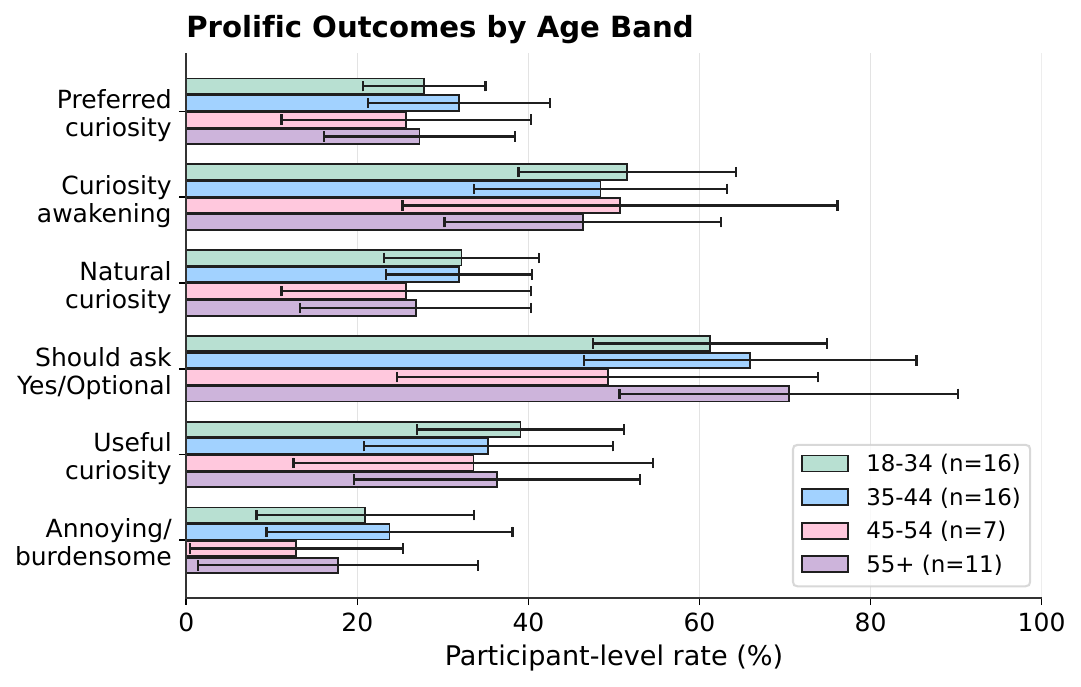}
\caption{Age-Based Differences in Prolific Study for Q3}
\label{fig:q3-age}
\end{figure}

We do not observe substantial differences across demographics, also because we have a small sample. Fig~\ref{fig:q3-gender} show that males are slightly higher than females on useful curiosity, curiosity awakening and natural curiosity while females are slightly higher on annoyance/burden. Fig~\ref{fig:q3-age} shows that age differences are small and noisy; the clearest visible difference is lower ``should ask a follow-up'' for 45-54 age group. However, we do not want to overclaim these differences due to the smaller sample size per demographic group.

\subsection{User Curiosity Modeling}
\label{app:useful_modeling}

\subsubsection{Regression Analysis}

We model useful curiosity at the item level. Each item consists of the original user prompt and the candidate curiosity-oriented follow-up question shown as Choice C in the human study. The prediction target is the item-level consensus label: \textit{validated useful} versus \textit{validated not useful}; ambiguous items are excluded from the classifier analysis.

For each non-ambiguous item \(i\), we define the binary target

  \[
  Y_i =
  \begin{cases}
  1 & \text{if item } i \text{ is validated useful},\\
  0 & \text{if item } i \text{ is validated not useful}.
  \end{cases}
  \]
  
We estimate logistic-regression models of the form
  \[
  \Pr(Y_i = 1 \mid X_i) = \sigma(\alpha + \beta^\top X_i),
  \]
where \(\sigma(\cdot)\) is the logistic link function and \(X_i\) is the feature representation for item \(i\). The learned coefficients \(\beta\) indicate which features increase or decrease the predicted probability that a curiosity-oriented question will be validated as useful.

We compare three feature sets. The first is \textbf{user-context affordance}, a one-hot categorical representation of the opportunity created by the user prompt. These labels capture whether the prompt presents an advice/dilemma, creative or research-framing request, uncertainty/information gap, reasoning/claim, or learning/explanation context. The second is \textbf{curiosity subtype}, a one-hot categorical representation of the taxonomy label assigned to the candidate curiosity question, including information-gap, causal-explanatory, exploratory-broadening, reflective/metacognitive, perspective-taking, and counterfactual/hypothetical. The third is \textbf{text features}, a TF-IDF representation computed from the concatenation of the user prompt and candidate curiosity question. Thus, the first two feature sets test interpretable categorical signals, while the TF-IDF model tests whether surface lexical information in the prompt and question provides predictive signal.

For all feature sets, we train logistic-regression classifiers to predict whether the candidate curiosity question is validated as useful. We compare against a majority-class baseline that always predicts the more frequent class, \textit{validated not useful}. Because the validated dataset is small, these results are intended as a proof of concept rather than a final predictive benchmark.

\subsection{Classifier stability}
The useful-curiosity classifier is intentionally lightweight and should not be interpreted as a deployable predictor. The validated dataset contains 57 non-ambiguous items, with 18 positive and 39 negative labels. We therefore report macro-F1 and AUC rather than accuracy alone, since the majority baseline obtains high accuracy by always predicting validated-not-useful. Because each cross-validation fold contains only a few positive examples, differences between feature sets should be interpreted qualitatively. The main takeaway is not that one model is final, but that user-context affordance and lexical context provide signal beyond a majority baseline.

\subsection{Prompting Pilot for Curiosity-Oriented Questioning}
\label{app:prompting_pilot}

To examine whether curiosity-oriented questioning
can be elicited through prompting, we conduct a
small pilot using \texttt{gpt-5-mini} on 80
missed-opportunity items from RQ3. We compare
three prompting conditions: direct answering,
asking questions when needed, and explicitly
encouraging curiosity-oriented questioning.
We classify the generated responses using
our question taxonomy.

\begin{table}[t]
\centering
\scriptsize
\begin{tabular}{lrrrr}
\hline
Prompt & Any Q & Curious Q & Other Q & Generic Q \\
\hline
Direct & 7.5 & 1.2 & 6.2 & 7.5 \\
Ask-if-needed & 96.2 & 52.5 & 43.8 & 41.2 \\
Curiosity & 97.5 & 83.8 & 11.2 & 11.2 \\
\hline
\end{tabular}
\caption{Percentage of generated responses classified
by question type in the 80-item prompting pilot.
Results are preliminary and do not measure
downstream user outcomes.}
\label{tab:curiosity_prompting_pilot}
\end{table}

As shown in Table~\ref{tab:curiosity_prompting_pilot},
the percentage of outputs containing curiosity-oriented
questions increases from 1.2\% under direct
answering to 52.5\% with ask-if-needed prompting
and 83.8\% with curiosity prompting. These
preliminary findings suggest that prompting can
substantially change the frequency of curiosity-oriented
questions. However, they do not establish whether
the generated questions improve downstream user
engagement or perceived usefulness.

\section{Use of AI assistants}
AI assistants were used minimally for language polishing and formatting support. All research ideas, experimental design, and final writing decisions were made by the authors.

\section{Reprodicibility}
We provide a lightweight reproducibility bundle for the Q1/Q2 analyses, including paper-facing scripts, sanitized sample/input data, and derived summary tables. The shared files remove raw user text, assistant text, model names, conversation identifiers, country fields, and user-level identifiers. We do not redistribute raw public conversation dumps or Prolific human-study data because they may contain user-generated text.



\end{document}